\documentclass[table]{gtech}
\PassOptionsToPackage{table, usenames, dvipsnames}{xcolor}
\usepackage{xcolor}
\usepackage{amssymb}
\usepackage{multirow}
\usepackage{bigdelim}
\usepackage{longtable}
\usepackage{tabularray}
\usepackage{wrapfig}
\usepackage[most]{tcolorbox}
\usepackage{url}
\usepackage{float}
\usepackage{datatool}
\usepackage{enumitem}
\usepackage{subcaption} 
\usepackage{caption}

\RequirePackage{tgpagella} 
\RequirePackage{mathpazo}  
\RequirePackage{inconsolata} 
\usepackage{makecell}
\usepackage{hyperref}
\usepackage{url}
\usepackage{times}
\usepackage{latexsym}
\usepackage[T1]{fontenc}
\usepackage[utf8]{inputenc}
\usepackage{microtype}
\usepackage{inconsolata}
\usepackage{amsmath}
\usepackage{amssymb}
\usepackage{xspace}
\usepackage{booktabs}
\usepackage{multirow}
\usepackage{wrapfig}
\usepackage{graphicx}
\usepackage{tabularx}
\usepackage{subcaption}
\usepackage{tcolorbox}
\usepackage{algorithm}
\usepackage{algpseudocode}
\usepackage{caption}
\usepackage{listings}

\newcommand{\paratitle}[1]{\vspace{1.5ex}\noindent\textbf{#1}}
\newcommand{\ie}{\emph{i.e.,}\xspace}

\newcommand{\eg}{\emph{e.g.,}\xspace}

\newcommand{\ignore}[1]{}
\newcommand{\OURS}{Ring-2.5-1T-Zero\xspace}
\newcommand{\OURSI}{Ring-2.5-1T-Zero-I\xspace}
\newcommand{\OURSII}{Ring-2.5-1T-Zero-II\xspace}
\newcommand{\OURSIII}{Ring-2.5-1T-Zero-III\xspace}

\newtcolorbox{promptbox}[2][Prompt]{%
  breakable, colback=#2!6!white, arc=5pt, boxrule=0.5pt,
  fonttitle=\bfseries, title=#1, colframe=#2!70!black,
  width=0.98\linewidth, left skip=0.01\linewidth, right skip=0.01\linewidth,
}

\algnewcommand{\Inputs}[1]{%
  \State \textbf{Inputs:}
  \Statex \hspace*{\algorithmicindent}\parbox[t]{0.98\linewidth}{\raggedright #1}
}
\algnewcommand{\Initialize}[1]{%
  \State \textbf{Initialize:}
  \Statex \hspace*{\algorithmicindent}\parbox[t]{0.98\linewidth}{\raggedright #1}
}

\renewcommand{\title}[1]{\newcommand{\titlelist}{{\huge\selectfont #1}}}

\title{\textbf{Ring-Zero: Scaling Zero RL to a Trillion Parameters for Emergent Reasoning}}

\author[1,*]{Xinyu Tang}
\author[2,*]{Qianggang Cao}
\author[1]{Yurou Liu}
\author[1]{Yuliang Zhan}
\author[3]{Xiaochong Lan}
\author[1]{Yifan Li}
\author[4]{Yuchen Yan}
\author[1]{Han Peng}
\author[1]{Zican Dong}
\author[2]{Zhenduo Zhang}
\author[2]{Tianshu Wang}
\author[2]{Xinyu Kong}
\author[2]{Zujie Wen}
\author[1,\ddag]{Wayne Xin Zhao}
\author[2,\ddag]{Zhiqiang Zhang}
\author[2]{Jun Zhou}

\renewcommand{\affiliationlist}{%
  {\normalsize $^{1}$Gaoling School of Artificial Intelligence, Renmin University of China\hspace{0.5em}$^{2}$Ant Group\\
  $^{3}$Tsinghua University\hspace{0.5em}$^{4}$Zhejiang University}%
}

\contribution[*]{Equal contribution.}
\contribution[\ddag]{Corresponding authors.}

\abstract{\fontsize{11pt}{12pt} 
Reinforcement learning with verifiable rewards~(RLVR) without human-annotated data, often referred to as ``zero RL'', has emerged as a powerful paradigm for eliciting chain-of-thought~(CoT) reasoning. However, due to computational constraints, existing studies are largely restricted to small models, leaving the training dynamics and emergent capabilities at a large scale unexplored.
To meaningfully explore this frontier, we aim to elicit high-quality reasoning behaviors from the model.
However, we find that naive scaling often suffers from poor readability, token redundancy, and a lack of adaptive reasoning depth. To address these challenges, we present a stable and efficient training pipeline, incorporating algorithmic and system optimizations such as clipped importance sampling, training-inference ratio correction, and mixed-precision control. 
Our experiments offer three key findings that validate the ``\textbf{bitter lesson}'' of scaling: (1) scaling to 1T parameters significantly enhances sample efficiency and performance ceilings; (2) the training process progresses sequentially through an initial ``discovery'' phase followed by a ``sharpening'' phase; and (3) the model spontaneously develops advanced cognitive behaviors, including anthropomorphism, structured formatting, self-verification, parallel reasoning, and context anxiety, rendering hand-crafted heuristics redundant. Evaluated on seven challenging mathematical benchmarks, \textbf{\OURS} achieves competitive performance. Additionally, to assess CoT quality beyond final-answer correctness, we propose a structured evaluation framework across three dimensions: comprehensibility, reproducibility, and efficiency, under which our model demonstrates clear advantages in producing structured and concise reasoning traces.
By sharing our experimental details and observed emergent phenomena, we hope to provide the community with deeper insights into scaling behaviors, particularly at the 1-trillion scale.
}

\addsecondlogo[3.8cm]{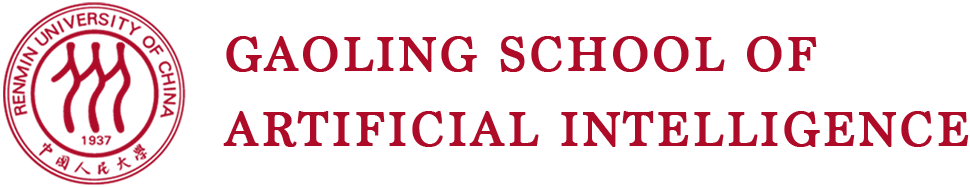} 
\begin{document}
\maketitle

\section{Introduction}
\label{sec:intro}

Alongside the scaling of model parameters, chain-of-thought~(CoT) reasoning~\citep{CoT-Wei-NIPS-2022,CoT-Kojima-NIPS-2022,Verify-Lightman-ICLR-2024} has emerged as a critical scaling dimension for test-time compute, establishing itself as a foundational capability for large language models~(LLMs)~\citep{LLM-survey,GPT3,Scaling-law} to solve complex tasks.
Recent work~\citep{GRPO} has demonstrated that reinforcement learning with verifiable rewards~(RLVR) can effectively enhance the reasoning abilities. 
In particular, the paradigm of ``zero RL''~\citep{DeepSeek-R1} diverges from traditional imitation learning by initiating RL directly from a pretrained base model, entirely bypassing the need for supervised CoT data, which has proven capable of incentivizing emergent problem-solving strategies.
This breakthrough indicates that powerful reasoning capabilities can be cultivated purely through trial-and-error reinforcement learning, circumventing the dependency on costly and curated reasoning demonstrations.

Over the past year, the research community has shown considerable interest in eliciting reasoning behaviors directly from base models. 
Extensive efforts have focused on high-quality data curation, generating diverse chain-of-thought datasets to bootstrap reasoning capabilities~\citep{Math-Data1,Math-Data2,Math-Data3}.
Building upon these foundations, a variety of policy optimization algorithms have been proposed to stabilize~\citep{GSPO,TIS,A3PO} and enhance optimization efficacy~\citep{DAPO,CISPO,Passk}.
To accommodate these intensive algorithmic workloads, scalable training infrastructures have also undergone a parallel evolution~\citep{OpenRLHF,Verl,slime,AReal}.
However, constrained by computational demands, the vast majority of existing studies rely on relatively small models.
While these studies provide valuable insights, a fundamental question remains unanswered: \emph{how do the training dynamics and emergent capabilities of zero RL evolve when applied to a trillion-parameter model?} 
Larger models possess substantially richer pretrained knowledge and latent capabilities that are amenable to optimization~\citep{DeepSeek-R1}. 
Exploring zero RL at this massive scale is therefore essential to charting the true boundaries of self-evolved reasoning.

To meaningfully explore this frontier, our primary goal is to elicit high-quality reasoning behaviors from the model, which naturally requires us to first define what makes a high-quality reasoning process.
While reaching the correct final answer is essential, the quality of the reasoning path itself is equally important. We believe an ideal CoT trace should not only be correct but also be readable to humans for easy verification, and computationally concise to save inference resources.
However, we find that a naive application of zero RL severely deviates from these ideals, exposing several critical limitations.
First, existing reasoning traces often suffer from \textbf{poor readability}. They typically lack logical formatting and clear structure, which makes it hard for humans to follow and interpret the thought process.
Second, standard algorithms like GRPO often introduce an implicit length bias by assigning disproportionate credit to longer sequences.
While token-level loss encourages exploration in early training, it eventually causes uncontrolled length growth and token redundancy, rendering the inference process highly \textbf{inefficient}.
Furthermore, different problems inherently require \textbf{varying levels of reasoning depth}, yet standard training pipelines typically produce a single-mode model constrained by a fixed response budget. 

In this work, we present a trillion-parameter self-iteration pipeline that trains models for emergent reasoning from scratch. 
A key insight from our exploration is that stabilizing training at the 1T scale only requires a few \textbf{simple yet critical} modifications.
Guided by this principle, our pipeline avoids heavy engineering. 
It simply combines clipped importance ratio policy gradient with training-inference ratio correction for incentivizing reasoning, self-distillation for compression and stabilization, sample-level loss normalization for stable training, and tier-based adaptive training for flexible reasoning depth. We also implement infrastructure optimizations, including mixed-precision control and context parallelism for stable and efficient training.

Directly answering our central question, our trillion-parameter experiments offer a profound validation of the ``\textbf{bitter lesson}'' in artificial intelligence: massive computation and scale ultimately overtake human-engineered heuristics. 
We summarize our insights into three key findings:
First, \textbf{scaling dictates the capability ceiling}. The 1T model exhibits vastly superior sample efficiency and reaches a significantly higher performance bound compared to its 104B counterpart.
Second, our analysis sheds light on the ``discovery vs. sharpening'' debate, suggesting that \textbf{both phenomena coexist as sequential stages}.
An initial ``discovery'' phase unlocks dormant pathways to expand the reasoning boundary, followed by a prolonged ``sharpening'' phase where the model refines its policy within this newly established boundary.
Third, and most strikingly, we observe the \textbf{spontaneous emergence of advanced cognitive strategies}.
While researchers historically engineered complex pipelines to incentivize improved thinking mechanisms, \OURS renders these hand-crafted heuristics redundant.
Driven purely by this self-iterative training process without any human-annotated data, the model autonomously converges on several key emergent behaviors as optimal mathematical solutions: \emph{anthropomorphism}, \emph{structured formatting}, \emph{self-verification}, \emph{parallel reasoning}, and \emph{context anxiety}.

Finally, we evaluate \OURS on seven challenging mathematical benchmarks and find that our model achieves competitive performance comparable to frontier models.
Furthermore, to assess CoT quality beyond mere final-answer accuracy, we propose a structured evaluation framework spanning three dimensions: \emph{comprehensibility}, \emph{reproducibility}, and \emph{efficiency}. 
Under these metrics, our model demonstrates clear advantages in generating structured, human-readable, and concise reasoning traces. 

Our main contributions are summarized as follows:

$\bullet$ We demonstrate that with only minor algorithmic and system improvements (\eg clipped importance sampling, training-inference ratio correction, and mixed-precision control), our self-iterative training process remains stable and efficient, successfully training \textbf{\OURS} to achieve competitive performance.

$\bullet$ We introduce a CoT quality evaluation framework that assesses reasoning traces along three dimensions: comprehensibility, reproducibility, and efficiency. This framework goes beyond final-answer accuracy and provides diagnostic signals for understanding the quality of the reasoning process itself.

$\bullet$ We provide an empirical validation of the bitter lesson at scale, demonstrating that scaling to a trillion-parameter model unlocks higher sample efficiency and performance ceilings while enabling the autonomous emergence of advanced cognitive strategies, including anthropomorphism, structured formatting, self-verification, parallel reasoning, and context anxiety, which renders hand-crafted designs redundant.
\section{Evaluation Metrics for Chain-of-Thought Quality}
\label{sec-metrics}

Evaluating the quality of Chain-of-Thought~(CoT) reasoning remains an open challenge.
Currently, most evaluations~\citep{GLM-5,Kimi-k2.5} rely on final-answer accuracy as the sole metric. 
However, accuracy only measures the final outcome, treating the underlying reasoning process as a black box. 
As LLMs are increasingly deployed in complex tasks and agentic workflows, CoT traces serve a dual purpose: they provide human-readable explanations and act as critical intermediate steps for downstream agent systems.
Therefore, a comprehensive evaluation must go beyond mere correctness.

Prior to our experiments, we pre-defined several essential characteristics that an ideal CoT trace should possess. 
Guided by these expectations, we construct a structured evaluation framework that assesses CoT quality along three complementary dimensions: \emph{Comprehensibility}, \emph{Reproducibility}, and \emph{Efficiency}.
Comprehensibility ensures the logic is transparent and easy for humans to read; Reproducibility measures whether the reasoning strategy can be effectively and efficiently learned by weaker models; and Efficiency evaluates whether the generation is concise and computationally cost-effective.
These three dimensions are independent of, yet highly complementary to, the correctness of the final answer.

\subsection{Comprehensibility}

Comprehensibility measures the degree to which a human reader can follow the reasoning steps without requiring external context or experiencing high cognitive load. 
A transparent CoT trace should feature a coherent logical flow, explicit causal dependencies between steps, and no hallucinated claims.
When any of these properties are violated, the trace fails its primary explanatory purpose, undermining user trust.

To evaluate comprehensibility quantitatively, we employ an \emph{LLM-as-a-Judge} framework via pairwise comparison. Given the same problem, a judge LLM evaluates CoT traces from different models and selects the one with stronger logical coherence, clearer articulation, and fewer reasoning flaws.
The complete evaluation prompt is detailed in Appendix~\ref{sec:eval_prompts}.

\subsection{Reproducibility}

Reproducibility assesses whether an agent lacking prior domain knowledge (\eg a human with novice or a weaker student model) can effectively and efficiently learn the underlying reasoning process and acquire comparable task-solving abilities.
An ideal trace encodes generalizable problem-solving strategies rather than intuitive leaps, enabling weaker student models to learn the underlying logic.

We measure reproducibility through \emph{knowledge distillation}. 
By fine-tuning a weaker model on the generated CoT traces, we use the downstream performance gain of the student as a proxy. 
A larger improvement indicates that the teacher's traces carry rich and transferable reasoning skills.

\subsection{Efficiency}

Efficiency evaluates the model's ability to solve a problem concisely, avoiding unnecessary verbosity, redundant derivations, or circular logic.
Furthermore, as CoT traces are heavily integrated into agent workflows, generation efficiency directly translates to faster inference speeds and lower token costs.
An efficient CoT isolates the essential path to the solution without sacrificing clarity or correctness.

In practice, we measure efficiency by calculating the \emph{average token count of correct CoT traces}. Among valid solutions, fewer tokens indicate higher efficiency, demonstrating that the model has identified a direct and optimal path to the answer rather than meandering through unrelated explorations.
\section{Methodology}
\label{sec:method}

\begin{figure*}[t]
    \centering
    \includegraphics[width=\textwidth]{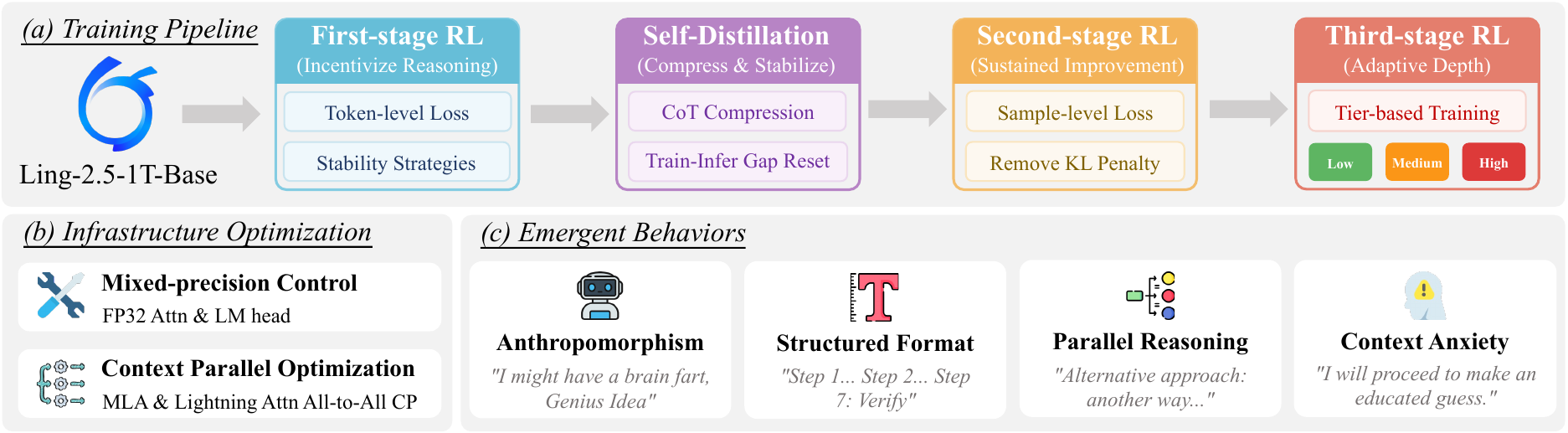}
    \caption{Overview of \OURS. (a) The multi-stage training pipeline. First-stage RL incentivizes reasoning from the base model. Self-Distillation compresses CoT traces and resets the training-inference engine gap. Second-stage RL shifts to a sample-level loss for sustained improvement. Third-stage RL introduces tier-based training for adaptive reasoning depth. (b) Infrastructure optimizations for stable and efficient training at scale. (c) Emergent behaviors that arise spontaneously without explicit supervision.}
    \label{fig:main}
\end{figure*}

In this section, we present a training framework that elicits and scales reasoning capabilities directly from a base model without relying on human-annotated data.
A core design principle of our framework is \textbf{minimalism}. Guided by our finding that overly complex algorithmic designs are often unnecessary, we only introduce optimization elements when they are strictly required to guide the model's iterative learning.
We first outline the multi-stage training pipeline (Section~\ref{sec:zero-rl}), which progressively builds reasoning capabilities through three RL stages and an intermediate self-distillation phase.
Subsequently, we detail the infrastructure optimizations (Section~\ref{sec:infra-optim}) that make long-context RL stable and efficient at scale.

\subsection{Zero Reinforcement Learning Training Pipeline}
\label{sec:zero-rl}

Our training pipeline consists of four distinct phases.
In the first RL stage (Section~\ref{sec:stage1}), we use a clipped importance ratio policy gradient, combined with KL regularization and a token-level loss, to stimulate reasoning from the base pretrained model.
Next, we apply self-distillation (Section~\ref{sec:self-distill}) to compress the learned reasoning traces and bridge the gap between training and inference.
The second RL stage (Section~\ref{sec:stage2}) resumes RL training using sample-level loss normalization for stable optimization.
Finally, the third RL stage (Section~\ref{sec:stage3}) introduces tier-based training, equipping the model with adaptive response modes for different difficulty levels.

\subsubsection{First Stage RL: Reasoning Elicitation}
\label{sec:stage1}

In Zero-RL, we train directly from a pretrained base model that lacks initial reasoning capabilities.
Such a model rarely generates step-by-step derivations because reasoning tokens have very low probabilities under the base policy.
To elicit these latent capabilities, the initial RL phase must aggressively amplify low-probability reasoning tokens.
Therefore, we adopt a \textbf{clipped importance-sampling policy gradient}~\citep{CISPO}.
Unlike standard PPO-clip, which entirely zeroes out gradients for tokens outside the clipping range, this approach applies a stop-gradient solely to the clipped ratio while allowing gradients to flow through for all tokens. 
This ensures every generated token contributes to the learning process, which is crucial for building reasoning abilities from scratch.
Specifically, the objective is formulated as:
\begin{equation}
\mathcal{J}(\theta) = \mathbb{E}_{\substack{q \sim \mathcal{D} \\ \{o_i\}_{i=1}^G \sim \pi_{\textcolor{red}{\text{S}}}}} \left[ \sum_{i=1}^{G} \sum_{t=1}^{|o_i|} \operatorname{sg}\!\left(\hat{\rho}_{i,t}\right) \cdot \hat{A}_{i,t} \cdot \log \pi_{\textcolor{blue}{\text{M}}}^{\theta}(o_{i,t} \mid q, o_{i,<t}) \right],
\label{eq:pg-obj}
\end{equation}
where $q$ is the input question sampled from dataset $\mathcal{D}$, $\{o_i\}_{i=1}^G$ are $G$ rollout responses from the inference engine $\pi_{\textcolor{red}{\text{S}}}$, $o_{i,t}$ is the $t$-th token of the $i$-th response, $\hat{A}_{i,t}$ is the advantage estimate computed using GRPO-style group-normalized rewards~\citep{DeepSeek-R1}, $\operatorname{sg}(\cdot)$ denotes the stop-gradient operator, and $\hat{\rho}_{i,t}$ is the clipped importance ratio defined below.

However, amplifying low-probability tokens amplifies a critical system-level vulnerability: the numerical discrepancy between the training engine (\textcolor{blue}{Megatron}) and the inference engine (\textcolor{red}{SGLang}). 
Differences in floating-point precisions and kernel implementations yield slightly divergent logits. 
To prevent these micro-discrepancies from compounding into macroscopic training collapse, we replace the numerator of the importance sampling ratio with the \textbf{actual training-engine logits}:
\begin{align}
\rho_{i,t} &= \frac{\pi_{\textcolor{blue}{\text{M}}}^{\theta}(o_{i,t} \mid q, o_{i,<t})}{\pi_{\textcolor{red}{\text{S}}}^{\theta_{\text{old}}}(o_{i,t} \mid q, o_{i,<t})}, \label{eq:policy-ratio} \\
\hat{\rho}_{i,t} &= \operatorname{clip}\!\left(\rho_{i,t},\; \epsilon_{\text{low}},\; \epsilon_{\text{high}}\right), \label{eq:clip}
\end{align}
where $\pi_{\textcolor{blue}{\text{M}}}^{\theta}$ is the current policy parameterized by $\theta$ in the training engine, $\pi_{\textcolor{red}{\text{S}}}^{\theta_{\text{old}}}$ is the rollout policy from the inference engine using previous parameters $\theta_{\text{old}}$, and $\hat{\rho}_{i,t}$ is the clipped ratio.
We do not apply a lower bound on the IS weight. 
Instead, we only set an upper bound $\epsilon_{\text{high}}$ to prevent excessively large updates.
This computed ratio accurately reflects the true divergence between training and inference engines.
To further stabilize training, we apply a \textbf{KL divergence penalty} against the frozen reference model, which prevents the policy from drifting too far from the reference model:
\begin{equation}
\mathcal{L}_{\text{KL}}(\theta) = \sum_{i=1}^{G} \sum_{t=1}^{|o_i|} D_{\text{KL}}\!\left(\pi_{\textcolor{blue}{\text{M}}}^{\theta}(\cdot \mid q, o_{i,<t}) \;\|\; \pi_{\text{ref}}(\cdot \mid q, o_{i,<t})\right),
\label{eq:kl}
\end{equation}
where $\pi_{\text{ref}}$ is the frozen reference policy model.

Finally, because base models tend to produce short responses, we explicitly encourage longer generation by adopting a \textbf{token-level loss} that sums over all tokens without normalizing by the response length $|o_i|$, as shown in Equation~\ref{eq:pg-obj}.
This ensures that longer correct responses accumulate stronger gradient signals.
To progressively improve reasoning capabilities, we gradually expand the response window length in a curriculum manner throughout training.
The combined first-stage RL objective \OURSI is:
\begin{equation}
\mathcal{L}_{\text{\OURSI}}(\theta) = -\mathcal{J}(\theta) + \beta \cdot \mathcal{L}_{\text{KL}}(\theta)
\label{eq:total-loss}
\end{equation}
where $\beta$ is the KL penalty coefficient that balances exploration and stability.

\subsubsection{Self Distillation: Compression and Stabilization}
\label{sec:self-distill}

While the first-stage RL successfully elicits reasoning capabilities, the unconstrained token-level loss introduces two critical drawbacks. 
First, the generated Chain-of-Thought (CoT) traces become excessively long, often burdened with redundant steps, circular logic, or unnecessary verbosity. 
Second, this extreme length severely destabilizes the RL process. 
Because the numerical discrepancies between the training and inference engines accumulate through the training process, it drastically inflates the training-inference gap, leading to optimization collapse.

To resolve these dual issues of verbosity and instability before proceeding to the next RL stage, we introduce a self-distillation phase to reset the model. 
Specifically, we use the first-stage expert policy $\pi_{\text{expert}}$ to sample multiple rollouts for each query. 
We then apply a two-step \textbf{length refinement process}: first, we select the shortest correct reasoning trace among these rollouts; second, we prompt the model to self-evaluate this selected trace and filter out any remaining segments it identifies as redundant. 
This mechanism effectively trims circular logic and compresses verbose derivations without sacrificing correctness.
Finally, this refined, high-quality corpus is used to fine-tune the original base model via standard supervised learning:
\begin{equation}
\mathcal{L}_{\text{self-distillation}}(\theta) = -\mathbb{E}_{q \sim \mathcal{D},\, o \sim \pi_{\text{expert}}} \left[ \sum_{t=1}^{|o|} \log \pi_{\theta}(o_t \mid q, o_{<t}) \right],
\label{eq:distill}
\end{equation}
where $o$ represents the length-refined response, and $\pi_{\theta}$ is the base pretrained model.
By distilling with shortened responses, the model acquires the expert's reasoning abilities while producing much more concise traces and effectively wipes out the accumulated numerical errors between the engines, which provides a highly capable, concise, and stable foundation for the subsequent RL stages.

\subsubsection{Second Stage RL: Sustained Optimization}
\label{sec:stage2}

After self-distillation, the model exhibits a much smaller training-inference gap.
This provides a stable starting point for the next phase of RL.
To maintain stability, we keep the same clipped importance-sampling policy gradient and ratio correction used in the first stage.

The key change in this stage is our loss calculation strategy.
In the first stage, a token-level loss is necessary to encourage longer responses.
However, the model now naturally produces long chains of thought.
Continuing with a token-level loss would cause the output length to grow without bound, destroying the conciseness gained from self-distillation.
To control response length, we pivot to a \textbf{sample-level loss}, which keeps the gradient magnitude independent of the output length:
\begin{equation}
\mathcal{L}_{\text{\OURSII}}(\theta) = - \mathbb{E}_{\substack{q \sim \mathcal{D} \\ \{o_i\}_{i=1}^G \sim \pi}} \left[ \sum_{i=1}^{G} \frac{1}{|o_i|} \sum_{t=1}^{|o_i|} \operatorname{sg}\!\left(\hat{\rho}_{i,t}\right) \cdot \hat{A}_{i,t} \cdot \log \pi_{\theta}(o_{i,t} \mid q, o_{i,<t}) \right],
\label{eq:total-loss-stage2}
\end{equation}
where the only change from Equation~\ref{eq:pg-obj} is the $\frac{1}{|o_i|}$ normalization factor.
Additionally, we \textbf{remove the KL penalty} in this stage.
Since the distilled model already provides a strong starting point, a KL constraint would unnecessarily restrict further exploration.
This sample-level formulation, along with the removal of KL regularization, ensures stable optimization over extensive RL steps.

\subsubsection{Third Stage RL: Adaptive Reasoning Depth}
\label{sec:stage3}

After two stages of RL, the model demonstrates strong reasoning abilities but operates in a single mode with a fixed-length budget.
In real-world scenarios, different questions require different levels of reasoning depth.
Simple questions need short answers, while complex problems benefit from extended reasoning paths.
To cultivate cognitive routing, we equip the model with multiple response modes by introducing a \textbf{tier-based training strategy} in the third stage.

We partition the training questions into three difficulty tiers, each governed by a specific maximum token length and a corresponding system prompt $p_k$:
\begin{equation}
\mathcal{T} = \left\{ \mathcal{T}_{\text{l}},\; \mathcal{T}_{\text{m}},\; \mathcal{T}_{\text{h}} \right\},
\label{eq:tiers}
\end{equation}
where $\mathcal{T}_{\text{l}}$, $\mathcal{T}_{\text{m}}$, and $\mathcal{T}_{\text{h}}$ denote the low, medium, and high difficulty tiers.
The third stage RL objective \OURSIII incorporates these prompts:
\begin{equation}
\mathcal{L}_{\text{\OURSIII}}(\theta) = -\sum_{k \in \{\text{l}, \text{m}, \text{h}\}} \mathbb{E}_{\substack{q \sim \mathcal{D}_k \\ \{o_i\}_{i=1}^G \sim \pi(\cdot \mid p_k, q)}} \left[ \sum_{i=1}^{G} \frac{1}{|o_i|} \sum_{t=1}^{|o_i|} \operatorname{sg}\!\left(\hat{\rho}_{i,t}\right) \cdot \hat{A}_{i,t} \cdot \log \pi_{\theta}(o_{i,t} \mid p_k, q, o_{i,<t}) \right],
\label{eq:pg-obj-stage3}
\end{equation}
where $\mathcal{D}_k$ is the subset of questions assigned to tier $k$, and $p_k$ is the corresponding system prompt.
The rollout policy $\pi$ generates responses conditioned on both the system prompt and the question, truncating the output at the given window size.
This mechanism teaches the model adaptive compute allocation: it learns to respond concisely to trivial queries while reserving exhaustive CoT generation for complex problems. 
During deployment, users can explicitly control this behavior via the system prompt.

\subsection{Infrastructure Optimization}
\label{sec:infra-optim}

Scaling RL to long context windows exposes critical bottlenecks in both numerical stability and communication latency. 
We address these challenges through two targeted optimizations.
First, we use mixed-precision control to ensure training stability (Section~\ref{sec:precision}).
Second, we optimize context parallelism to maximize throughput (Section~\ref{sec:cp-optim}).

\subsubsection{Mixed-precision Control}
\label{sec:precision}

During RL training, the importance ratio $\rho_{i,t}$ computes the ratio of two token probabilities.
These probabilities are derived from logits using a softmax function, which involves exponentiation.
The exponential function naturally amplifies small numerical errors.
A minor error in a logit can blow up into a massive gap in the probability ratio, completely destabilizing the training process. 
Consequently, any forward-pass component involving exponentiation is highly sensitive to precision. Specifically, this points to two key areas: the attention softmax and the LM head.

To solve this problem, we implement a simple yet effective mixed-precision strategy. 
We keep the main model body in BF16 but compute both the attention softmax and the LM head in FP32. 
By using high precision strictly in these two sensitive locations, we prevent small rounding errors from growing into large RL losses. 
In practice, this targeted adjustment eliminates sudden spikes in the loss and successfully closes the numerical gap between the training engine (\textcolor{blue}{Megatron}) and the inference engine (\textcolor{red}{SGLang}).

\subsubsection{Context parallelism Optimization}
\label{sec:cp-optim}

Training with long context windows requires context parallelism (CP) to partition a single long sequence across multiple devices.
Standard ring attention~\citep{Ring-attention} solves this by passing key-value (KV) chunks sequentially in a circle.
However, this design forces each device to wait for a complete round of communication with all other devices to finish its attention calculation. 
As the scale of parallelism grows, this sequential dependency creates a severe latency bottleneck.

To break this bottleneck, we tailor our CP strategy to our model's hybrid architecture, which consists of MLA and Lightning Attention layers. 
For the MLA layers~\citep{MLA}, we adopt an all-to-all CP strategy.
This approach reshuffles data along the head dimension using a single collective operation.
As a result, each device can independently compute full attention for a subset of heads across the entire sequence.
Because MLA compresses keys and values into a low-rank latent space, the all-to-all communication volume is drastically reduced compared to sending full KV tensors.
For the Lightning Attention layers~\citep{lightning-attention}, the calculation depends on a small, fixed-size KV state matrix.
Instead of passing this state sequentially around the ring, we use a single AllGather operation. This broadcasts all local states to every device at once, allowing them to finish their computations immediately.
These optimizations are mathematically equivalent to standard ring attention and produce identical gradients.
\section{Experiments}
\label{sec-experiments}

This section presents our experimental evaluation.
We first describe the experimental setup, then report the main results to compare our models against current frontier models, and finally provide a detailed analysis to validate our key design choices.

\subsection{Experimental Setup}

We summarize the key components of our experimental setup below. This covers model configurations, training details, reward design, and the evaluation metrics.

\paratitle{Base models.}
We conduct zero RL training starting from two pretrained Ling-2.5 base models~\citep{Ling-1T} without any supervised fine-tuning:
The models are:
(1)~\textbf{Ling-2.5-1T-Base}, a 1-Trillion parameter Mixture-of-Experts (MoE) model with 63B activated parameters.
(2)~\textbf{Ling-2.5-flash-Base}, a 104B parameter MoE model with 7.4B activated parameters.
Both models are trained from scratch using our four-stage pipeline. This progresses from First Stage RL to Self-Distillation, Second Stage RL, and finally Third Stage RL.

\paratitle{Training infrastructure.}
We run all experiments on 320 $\times$ H200 GPUs.
We adopt a hybrid architecture where \textcolor{blue}{Megatron} serves as the training engine and \textcolor{red}{SGLang} serves as the rollout engine.
The RL training process is orchestrated by the Areal~\citep{AReal} framework.

\paratitle{Hyperparameters.}
Our training first follows an off-policy RL setup with a batch size of 512 and a minibatch size of 32, then uses a batch size of 256 and a minibatch size of 32. 
Across all RL stages, we use the Adam optimizer with $\beta_1 = 0.9$, $\beta_2 = 0.999$, a constant learning rate of $2 \times 10^{-6}$, and a weight decay of $0.01$.
For each question, we generate $G = 16$ rollout responses at a temperature of $1.0$.
To encourage exploration, we do not impose a lower bound on the importance sampling ratio. We only clip the upper bound with $\epsilon_{\text{high}} = 5.0$ to prevent excessively large updates.
In the first stage RL, we use a token-level loss combined with a KL penalty ($\beta = 10^{-4}$, K3 divergence).
We update the reference model every 400 steps with the latest checkpoint. This ensures the KL anchor remains relevant as the policy improves.
We progressively expand the response window from 4k to 64k tokens in a curriculum method, doubling the context window every 800 training steps.
After the first stage RL, we curate and filter high-quality responses from the first stage RL expert model. We then fine-tune the base model for 3 epochs with a sequence length of 64k and a learning rate of $7 \times 10^{-5}$.
In the second stage RL, we switch to a sample-level loss and remove the KL penalty. This adjustment prevents uncontrolled length growth and enables highly stable training.
In third stage RL, we partition questions into Low (4k), Medium (16k), and High (64k) difficulty tiers. Each tier receives a specific system prompt. This teaches the model to adapt its reasoning depth based on the available inference budget.

\paratitle{Reward design.}
Throughout training, the reward consists of an accuracy component and a format component:
\begin{equation}
r_i = r_{\text{acc},i} + r_{\text{format},i},
\end{equation}
where $r_{\text{format},i} \in \{0, 1\}$ checks whether the response follows the required structure with \texttt{<think>...} \texttt{</think>} and \texttt{<answer>...</answer>} tags, and $r_{\text{acc},i} \in \{0, 1\}$ measures the final answer's correctness.
In early training stages, we use easily verifiable problems. Following previous work~\citep{DeepSeek-R1}, their correctness can be judged by deterministic rule-based matching.
In later stages, problems become significantly harder. Answers may take multiple valid forms or require semantic understanding. Therefore, rule-based verification becomes unreliable. 
We replace $r_{\text{acc},i}$ with an LLM-as-a-Judge evaluator (\ie Qwen3-Next-80B-A3B-Instruct). This judge receives both the model's response and the reference answer to output a binary correctness judgment.

\paratitle{Evaluation metrics.}
We evaluate our models on seven challenging mathematical reasoning benchmarks: AIME 2024, AIME 2025, AIME 2026, HMMT February 2025, HMMT November 2025, HMMT February 2026, and IMOAnswerBench.
Following ~\citet{DeepSeek-R1}, we set the temperature to 0.6 and top-k to 0.95 during inference.
For all evaluations, we report pass@1 accuracy averaged over 64 runs to minimize variance.
For our multi-tier model (Third Stage RL), we report results under three distinct inference modes: High (64k budget), Medium (16k budget), and Low (4k budget).

\subsection{Main Results}

\begin{figure*}[t]
    \centering
    \begin{subfigure}{0.24\textwidth}
        \centering
        \includegraphics[width=\linewidth]{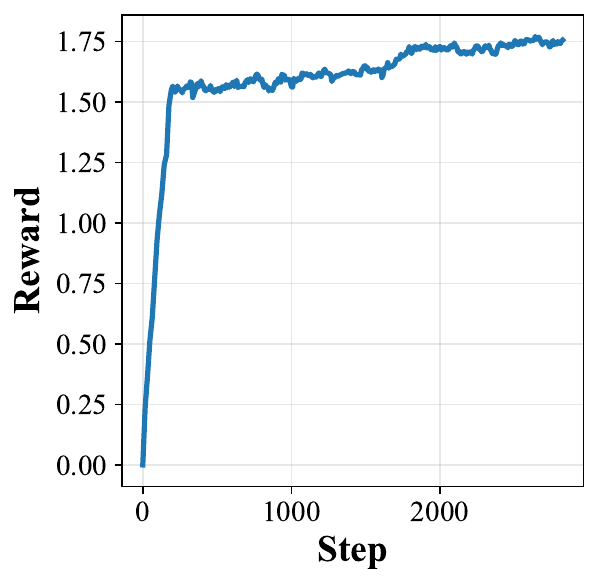}
        \caption{Reward (initial data)}
    \end{subfigure}
    \hfill
    \begin{subfigure}{0.24\textwidth}
        \centering
        \includegraphics[width=\linewidth]{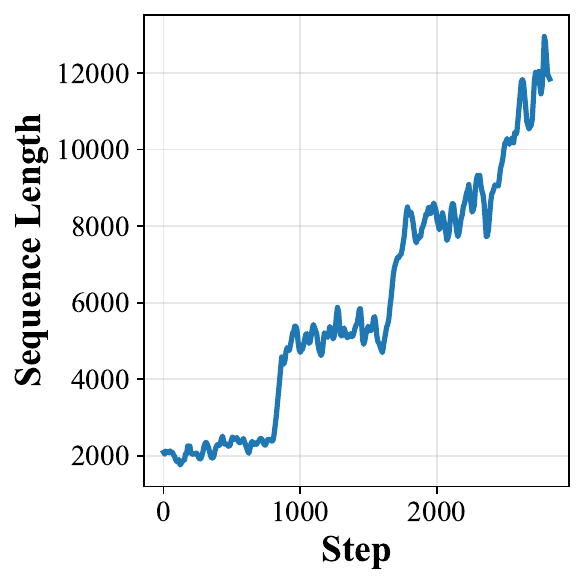}
        \caption{Seq Length (initial data)}
    \end{subfigure}
    \hfill
    \begin{subfigure}{0.24\textwidth}
        \centering
        \includegraphics[width=\linewidth]{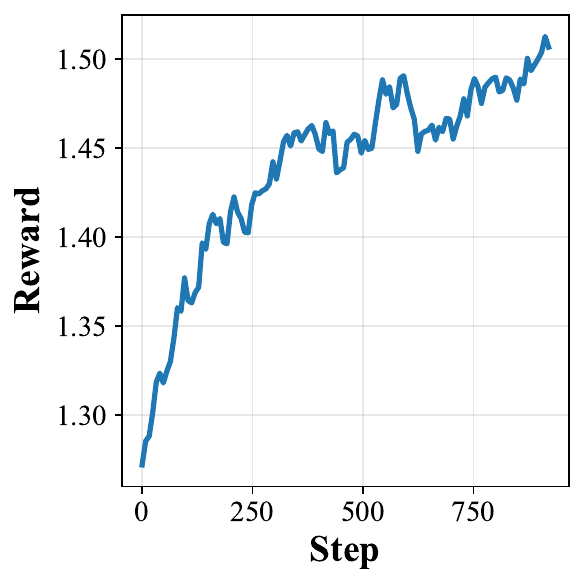}
        \caption{Reward (new data)}
    \end{subfigure}
    \hfill
    \begin{subfigure}{0.24\textwidth}
        \centering
        \includegraphics[width=\linewidth]{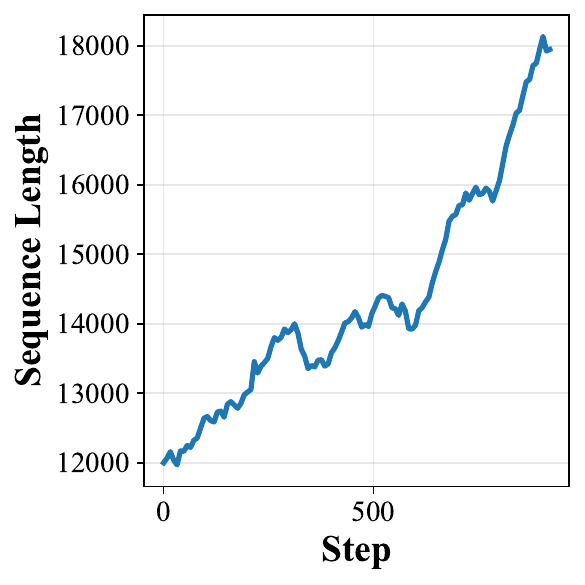}
        \caption{Seq Length (new data)}
    \end{subfigure}
    \caption{Training curves of Ling-2.5-1T-Base during first stage RL. (a,b)~First 2800 steps with the initial training data: reward and sequence length increase steadily as the model bootstraps reasoning from scratch. (c,d)~After switching to new training data, the model continues to improve with sustained sequence length growth.}
    \label{fig:train-max}
\end{figure*}

Table~\ref{tab:main-results} presents the main results comparing our \OURS models against state-of-the-art proprietary models across various mathematical reasoning benchmarks. 
Furthermore, Figure~\ref{fig:train-max} illustrates the comprehensive training dynamics of our First Stage RL. By dividing the training into two distinct phases with progressively difficult data (\ie starting with relatively simpler data followed by harder data to push the upper bounds of reasoning), we ensure a stable and continuous optimization process.

\begin{table*}[t]
\centering
\caption{Main results on mathematical reasoning benchmarks. All scores are pass@1 accuracy (\%). ``TT'' denotes the truncation window size during training, ``IT'' indicates the average number of tokens during inference.}
\label{tab:main-results}
\resizebox{\textwidth}{!}{
\begin{tabular}{lccccccc}
\toprule
\textbf{Model} & 
\begin{tabular}[c]{@{}c@{}}\textbf{AIME} \\ \textbf{2024}\end{tabular} & 
\begin{tabular}[c]{@{}c@{}}\textbf{AIME} \\ \textbf{2025}\end{tabular} & 
\begin{tabular}[c]{@{}c@{}}\textbf{AIME} \\ \textbf{2026}\end{tabular} & 
\begin{tabular}[c]{@{}c@{}}\textbf{HMMT} \\ \textbf{Feb. 2025}\end{tabular} & 
\begin{tabular}[c]{@{}c@{}}\textbf{HMMT} \\ \textbf{Nov. 2025}\end{tabular} & 
\begin{tabular}[c]{@{}c@{}}\textbf{HMMT} \\ \textbf{Feb. 2026}\end{tabular} & 
\begin{tabular}[c]{@{}c@{}}\textbf{IMOAnswer} \\ \textbf{Bench}\end{tabular} \\ 
\midrule
\multicolumn{8}{c}{\multirow{1.5}{*}{\Large\textit{\textbf{Frontier Models}}}} \\
\multicolumn{8}{c}{} \\
GLM-5.1          & -    & -    & 95.3 & -    & 94.0 & 82.6 & 83.8 \\
DS-V4-Pro Max    & -    & -    & 94.6 & -    & 94.4 & 95.2 & 89.8 \\
Qwen3.7-Plus     & -    & -    & 97.0 & -    & 95.0 & 92.9 & 86.0 \\
Kimi K2.6        & -    & -    & 96.4 & -    & -    & 92.7 & 86.0 \\
Minimax M2.7     & -    & -    & 89.8 & -    & 81.0 & 72.7 & 66.3 \\
Claude Opus 4.8  & -    & -    & 95.7 & -    & 96.5 & 96.7 & 83.5 \\
Gemini 3.1 Pro   & -    & -    & 98.2 & -    & 94.8 & 87.3 & 81.0 \\
GPT-5.5          & -    & -    & 98.3 & -    & 96.5 & 96.7 & 91.4 \\
\midrule
\multicolumn{8}{c}{\multirow{1.5}{*}{\Large\textit{\textbf{Zero RL}}}} \\
\multicolumn{8}{c}{} \\
DeepSeek-R1-Zero                        & 77.9 & - & - & - & - & - & - \\
DeepSeek-R1                             & 79.8 & - & - & - & - & - & - \\
Ring-2.5-flash-Zero (First Stage RL)    & 71.2 & 63.5 & 65.3 & 55.2 & 54.8 & 50.3 & -    \\
Ring-2.5-flash-Distilled                & 86.2 & 77.2 & 78.0 & 70.3 & 68.5 & 62.6 & -    \\
\OURS (First Stage RL)                 & 89.1 & 83.3 & 84.2 & 76.7 & 75.8 & 66.2 & 59.3 \\
\OURS (Self Distillation)              & 92.3 & 87.3 & 88.1 & 81.9 & 79.9 & 71.2 & 63.8 \\
\OURS (Second Stage RL)                & 93.5 & 91.6 & 92.5 & 87.4 & 87.1 & 78.1 & 72.7 \\
\OURS (Second Stage RL, Yarn=2)        & 94.1 & 92.3 & 93.2 & 90.6 & 90.8 & 81.0 & 75.5 \\
\OURS (Third Stage RL) \\
\quad -- Low (TT=4k, IT=2353)             & 82.3 & 64.0 & 68.8 & 61.3 & 65.3 & 52.5 & 54.1   \\
\quad -- Medium (TT=16k, IT=8085)         & 90.9 & 88.1 & 90.8 & 80.4 & 84.8 & 74.1 & 70.8    \\
\quad -- High (TT=128k, Yarn=2, IT=20817) & 93.2  & 91.0 & 91.4 & 86.3 & 86.4 & 78.4 & 72.7  \\
\bottomrule
\end{tabular}
}
\end{table*}

\paratitle{Zero RL achieves competitive reasoning from scratch.}
Without relying on any human-annotated data, \OURS (First Stage RL) already achieves an impressive 84.2\% accuracy on AIME 2026. This compelling result underscores that our First Stage RL can effectively bootstrap advanced mathematical reasoning directly from a pretrained base model. 
The progressive curriculum, which gradually scales the response window from 4k to 64k tokens, combined with importance ratio correction, stably and effectively incentivizes the model to develop emergent chain-of-thought (CoT) behaviors. 
This suggests that as long as the exploration space is properly constrained via careful regularization, large language models inherently possess the capacity to self-discover complex logical deductions without human-annotated trajectories.

\paratitle{Multi-stage training yields consistent improvement.}
Each subsequent stage in our pipeline yields measurable performance gains, validating the overall efficacy of our multi-stage methodology.
Specifically, the self-distillation phase stabilizes the model and prepares it for extended RL optimization. Subsequently, the Second Stage RL incorporates a sample-level loss to facilitate sustained performance improvements, while the Third Stage RL applies tier-based training to embed adaptive reasoning depth into the model's capabilities.
By transitioning the granularity of supervision from token-level (First Stage RL) to sample-level (Second Stage RL), and using self-distillation as an anchor, we create a much smoother optimization landscape for sustained reasoning improvement.

\paratitle{Adaptive inference modes offer flexible performance-efficiency trade-offs.}
The three inference tiers of \OURS (\ie Low, Medium, and High) provide users with fine-grained control over reasoning depth during inference. 
While the Medium (TT=16k) and Low (TT=4k) modes effectively reduce inference latency and compute costs while maintaining competitive performance, we observe a slight performance drop in the Third Stage RL compared to the Second Stage peak. 
This decline is mainly due to two factors: first, a lack of high-quality, ultra-long reasoning data limits the High mode's upper bound. Second, jointly training across three different lengths introduces negative transfer, where the training signals from the Low and Medium modes slightly pull down the High mode's peak capability. 
Despite this, our adaptive strategy successfully shifts compute scaling from a rigid, one-size-fits-all approach to a dynamic, query-dependent allocation, proving that many queries simply do not require exhaustive reasoning.

\paratitle{Scaling model size amplifies Zero RL benefits.}
By comparing the 104-billion-parameter Ring-2.5-flash-Zero with the 1-trillion-parameter \OURS, we observe that larger model capacity disproportionately benefits from the Zero RL paradigm. The performance gap widens significantly on harder benchmarks.
This trend provides a clear demonstration that aggressive parameter scaling is of paramount importance, particularly in the Zero RL setting.
Under the zero RL regime, a model's ability to explore is strictly bounded by its internal world model. More parameters provide the necessary representational capacity and broader internal knowledge base required to autonomously navigate the massive search space.

\subsection{Other Evaluations of CoT Quality}

Beyond evaluating final-answer accuracy, we assess the quality of our model's CoT reasoning trajectories across three complementary dimensions: \emph{comprehensibility}, \emph{reproducibility}, and \emph{efficiency}. 
As demonstrated below, our zero-RL model not only achieves high accuracy but also produces reasoning traces that are highly readable, easily transferable, and remarkably token-efficient.

\subsubsection{Comprehensibility}

Beyond quantitative scaling, we observe an intriguing qualitative phenomenon: the 1T model spontaneously develops structurally segmented reasoning. Even without explicit formatting constraints, it naturally demarcates reasoning phases using markers like \texttt{Step 1:} and \texttt{Step 2:} (See Section~\ref{subsec:beautiful_format}).
Unlike the highly sanitized summaries typical of human reference solutions, these emergent traces capture an authentic, search-like trajectory characterized by trial-and-error, self-reflection, and parallel exploration.
The spontaneous emergence of these step markers offers a profound insight. It suggests that the model does not merely engage in a continuous, greedy generation stream; rather, it exhibits a form of \textbf{localized forward planning}, setting cognitive intents for the upcoming chunk of tokens before actually generating them.

To evaluate comprehensibility quantitatively, we conduct pairwise comparisons using an \emph{LLM-as-a-Judge} protocol. Across all 90 AIME problems (2024 to 2026), we compare our model's reasoning traces against four strong baselines (GLM-5.1, Kimi-k2.6, MiniMax-M2.7, and Qwen3.5-397B), focusing on logical coherence, causal explicitness, and the absence of hallucinations. 
As shown in Figure~\ref{fig:cot-quality}(a), our model achieves dominant win rates across all comparisons. This confirms that the emergent structured reasoning reflects a genuinely superior logical organization that human readers find easier to understand and follow.

\subsubsection{Reproducibility}

A key indicator of CoT quality is whether the reasoning traces can serve as an effective training signal for weaker models. We evaluate this reproducibility by distilling our model's reasoning traces into Qwen2.5-32B and Llama3.3-70B-Instruct models, and compare them against the same config distilled from DeepSeek-R1.

Notably, we achieve our distillation results using \textbf{only 100K} data samples, a mere fraction compared to the 800K samples utilized by DeepSeek-R1. 
As shown in Figure~\ref{fig:cot-quality}(b), despite using significantly less data, models distilled from our traces consistently outperform their DeepSeek-R1 counterparts. We observe a 5.8-point gain on Qwen-32B (78.4 vs.\ 72.6) and a 4.5-point gain on Llama-70B (74.5 vs.\ 70.0).
This demonstrates that our step-segmented reasoning traces provide a richer, more transferable learning signal, resulting in \textbf{vastly superior sample efficiency} during knowledge distillation.

\subsubsection{Efficiency}

Finally, we assess whether our model achieves high accuracy using fewer reasoning tokens. We identify a subset of problems where all five models (ours and the four baselines) answer correctly. We then compare the average token count of their CoT traces on this shared subset.

\begin{figure*}[t]
    \centering
    \begin{subfigure}{0.32\textwidth}
        \centering
        \includegraphics[width=\linewidth]{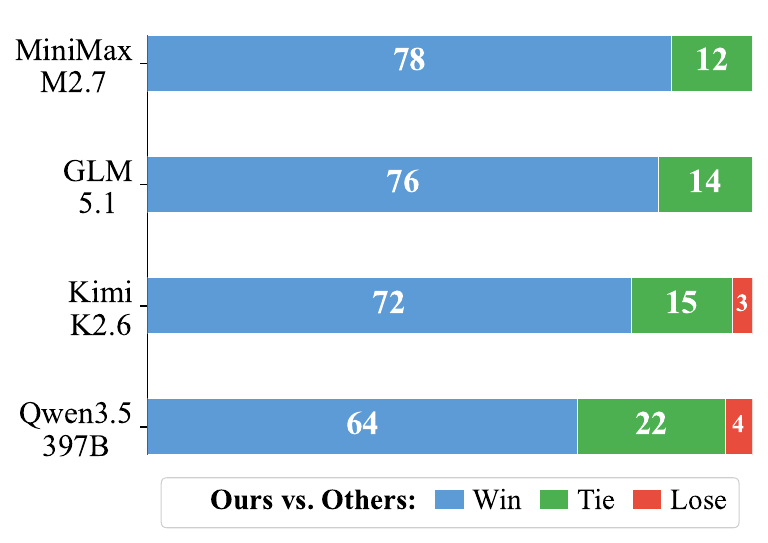}
        \caption{Comprehensibility}
    \end{subfigure}
    \hfill
    \begin{subfigure}{0.32\textwidth}
        \centering
        \includegraphics[width=\linewidth]{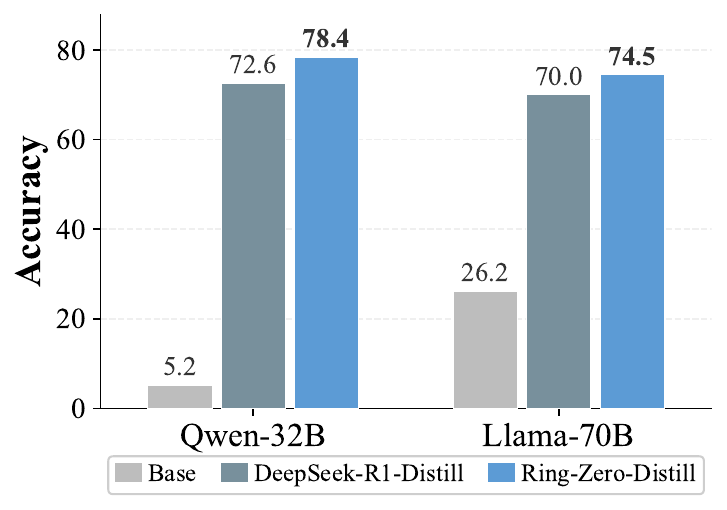}
        \caption{Reproducibility}
    \end{subfigure}
    \hfill
    \begin{subfigure}{0.32\textwidth}
        \centering
        \includegraphics[width=\linewidth]{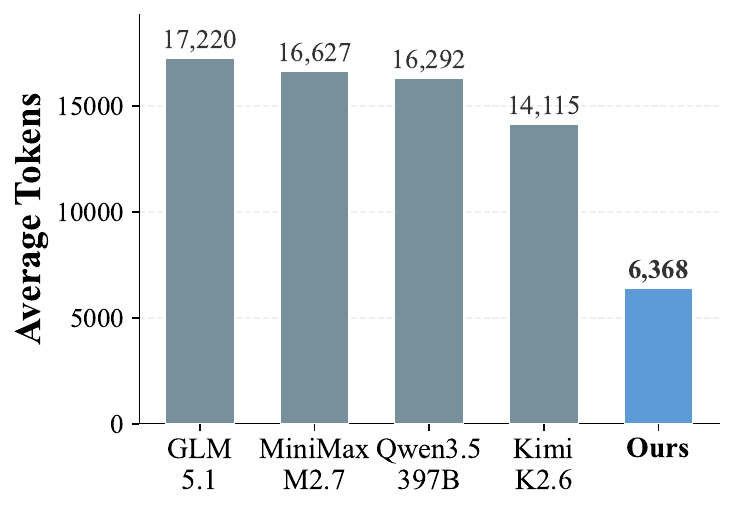}
        \caption{Efficiency}
    \end{subfigure}
    \caption{Evaluation of CoT quality across three dimensions. (a) Comprehensibility: our model's reasoning traces are judged to be more comprehensible than all baselines. (b) Reproducibility: distilling from our fewer CoT traces yields much stronger student models compared to DeepSeek-R1, highlighting a significantly higher sample efficiency for ability transfer. (c) Efficiency: our model solves problems using significantly fewer tokens.}
    \label{fig:cot-quality}
\end{figure*}

As shown in Figure~\ref{fig:cot-quality}(c), on the mutually solved AIME problems, our model uses an average of only 6,368 tokens. This is less than half of the other models. This substantial efficiency advantage suggests that our CoT helps the model eliminate redundant exploratory steps, allowing it to arrive at correct solutions via much more direct logical paths.

\subsection{Detailed Analysis}

In this subsection, we present detailed ablation studies to validate our key design choices. Due to computational resource constraints, we conduct all ablation experiments on the Ling-2.5-flash-Base model.

\subsubsection{The Impact of Different RL Algorithms}

\begin{figure*}[t]
    \centering
    \begin{subfigure}{0.24\textwidth}
        \centering
        \includegraphics[width=\linewidth]{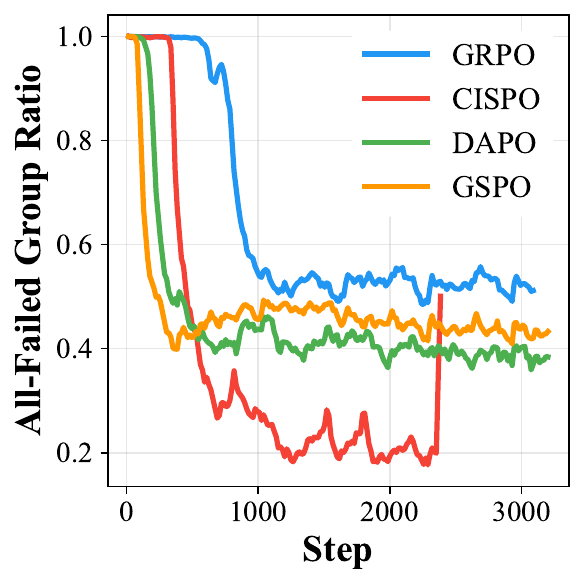}
        \caption{All-Failed Group Ratio}
    \end{subfigure}
    \hfill
    \begin{subfigure}{0.24\textwidth}
        \centering
        \includegraphics[width=\linewidth]{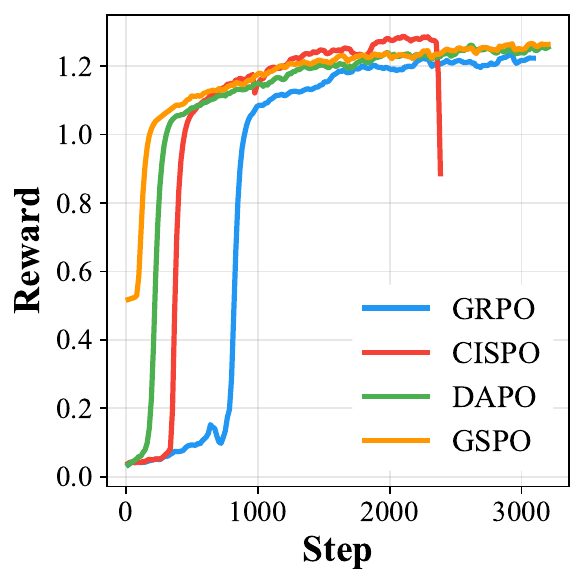}
        \caption{Reward}
    \end{subfigure}
    \hfill
    \begin{subfigure}{0.24\textwidth}
        \centering
        \includegraphics[width=\linewidth]{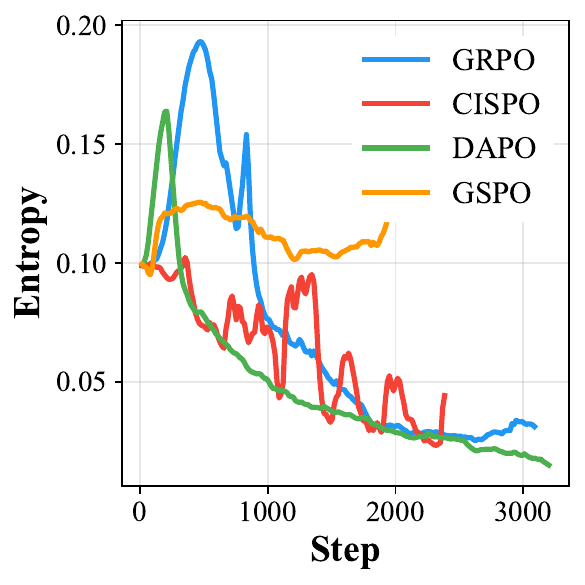}
        \caption{Entropy}
    \end{subfigure}
    \hfill
    \begin{subfigure}{0.24\textwidth}
        \centering
        \includegraphics[width=\linewidth]{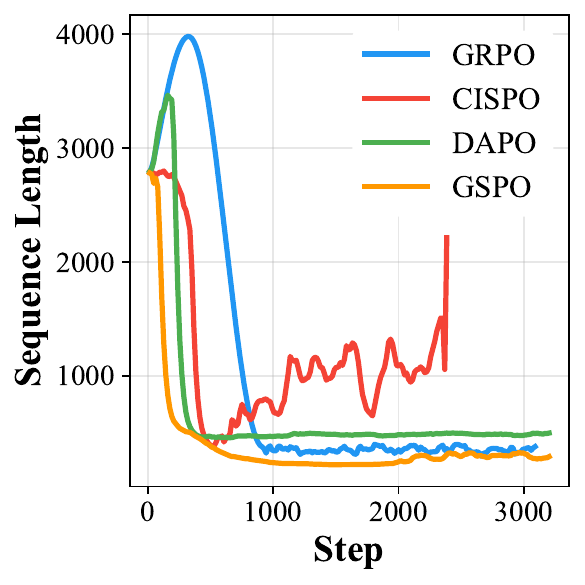}
        \caption{Sequence Length}
    \end{subfigure}
    \caption{Comparison of RL algorithms on the flash model. CISPO and DAPO accelerate learning but suffer from greater instability. GSPO maintains high entropy but provides limited sequence length growth.}
    \label{fig:rl-algos}
\end{figure*}

We compare four RL algorithms under identical settings: GRPO~\citep{DeepSeek-R1}, DAPO~\citep{DAPO}, CISPO~\citep{CISPO}, and GSPO~\citep{GSPO}.
All methods share the same base model, learning rate, and training data. They differ exclusively in how the policy gradient is computed.
As shown in Figure~\ref{fig:rl-algos}, both CISPO and DAPO build upon GRPO by amplifying the gradient signal for low-probability tokens. This effectively reduces the all-failed group ratio much faster than vanilla GRPO. Both algorithms also achieve faster reward growth and a more rapid increase in sequence length.
However, this acceleration comes at the cost of stability. CISPO is the most prone to entropy collapse and training instability, followed by DAPO and then GRPO.
The speed of reward improvement and the degree of instability follow the exact same ordering: CISPO $>$ DAPO $>$ GRPO.
This confirms our hypothesis: amplifying low-probability tokens is highly effective for stimulating reasoning from scratch, but it requires careful stabilization.
In contrast, GSPO uses a sample-level loss to explicitly maintain high entropy. While this prevents entropy collapse, it provides very little incentive for the model to explore longer sequence lengths. Consequently, it is ill-suited for bootstrapping early reasoning abilities.

\subsubsection{The Impact of RL Stabilization Strategies}

The analysis above reveals a fundamental trade-off. Algorithms that aggressively amplify low-probability tokens learn faster but are inherently less stable.
Therefore, we investigate two complementary stabilization strategies that preserve fast learning while maintaining training robustness.

\paratitle{KL penalty.}
\begin{figure*}[t]
    \centering
    \begin{subfigure}{0.24\textwidth}
        \centering
        \includegraphics[width=\linewidth]{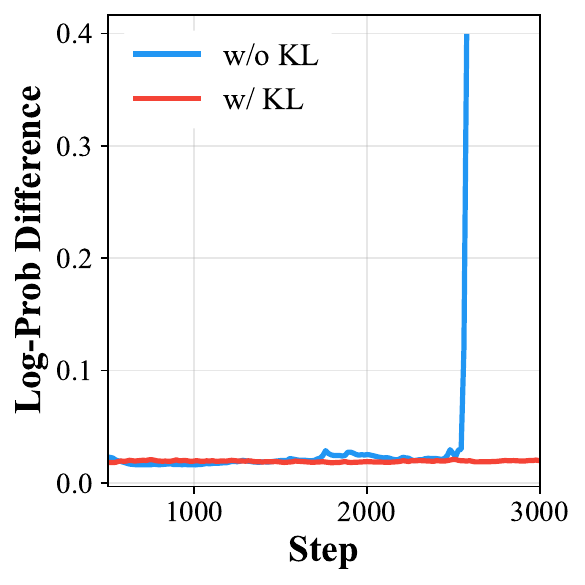}
        \caption{Log-Prob Difference}
    \end{subfigure}
    \hfill
    \begin{subfigure}{0.24\textwidth}
        \centering
        \includegraphics[width=\linewidth]{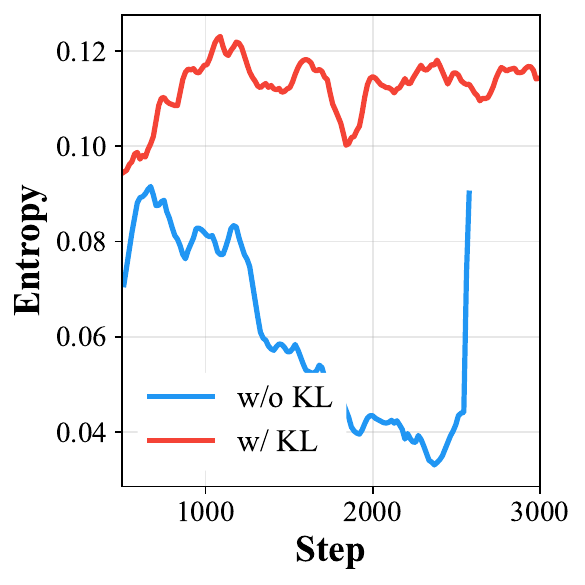}
        \caption{Entropy}
    \end{subfigure}
    \hfill
    \begin{subfigure}{0.24\textwidth}
        \centering
        \includegraphics[width=\linewidth]{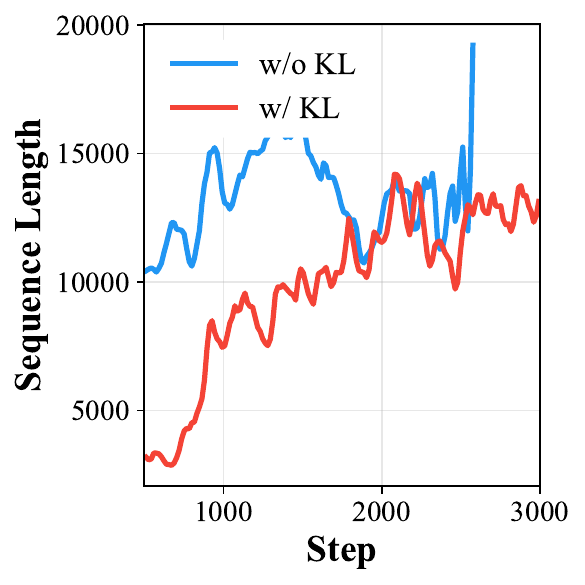}
        \caption{Sequence Length}
    \end{subfigure}
    \hfill
    \begin{subfigure}{0.24\textwidth}
        \centering
        \includegraphics[width=\linewidth]{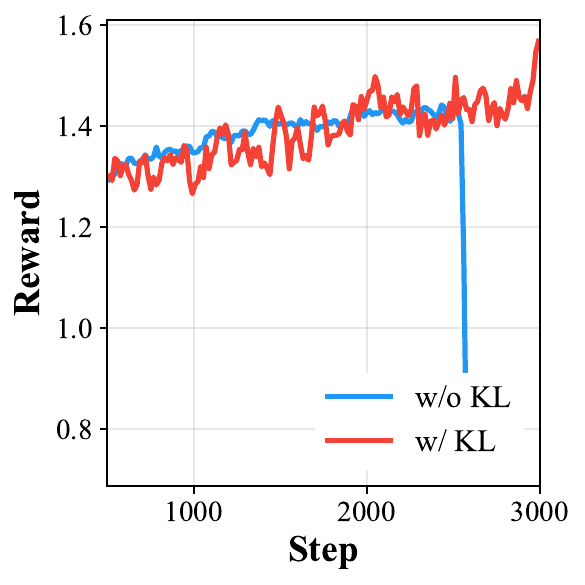}
        \caption{Reward}
    \end{subfigure}
    \caption{Effect of KL penalty on training stability. Without KL (blue), the training-inference log-probability gap diverges, causing the reward to crash. With KL (red), all metrics remain healthy.}
    \label{fig:kl-penalty}
\end{figure*}
We compare training runs with and without a KL divergence penalty against the reference model (Eq.~\ref{eq:kl}).
As shown in Figure~\ref{fig:kl-penalty}, removing the KL penalty causes catastrophic failure. The log-probability difference between the training and rollout engines grows unboundedly. Entropy collapses, and the reward eventually crashes.
With the KL penalty, all training metrics remain stable. The numerical gap stays controlled, entropy remains at a healthy level to preserve exploration, and the reward improves steadily.
The KL penalty acts as a vital regularizer. It constrains the overall policy drift, prevents premature entropy collapse, and sustains the model's exploratory capacity.

\begin{figure*}[t]
    \centering
    \begin{subfigure}{0.24\textwidth}
        \centering
        \includegraphics[width=\linewidth]{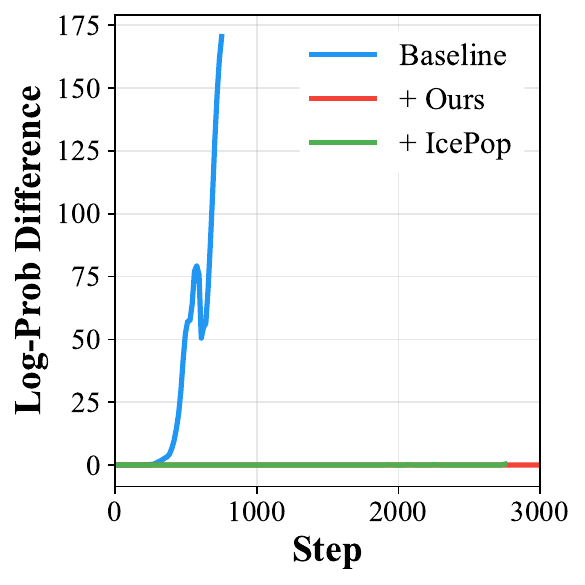}
        \caption{Log-Prob Difference}
    \end{subfigure}
    \hfill
    \begin{subfigure}{0.24\textwidth}
        \centering
        \includegraphics[width=\linewidth]{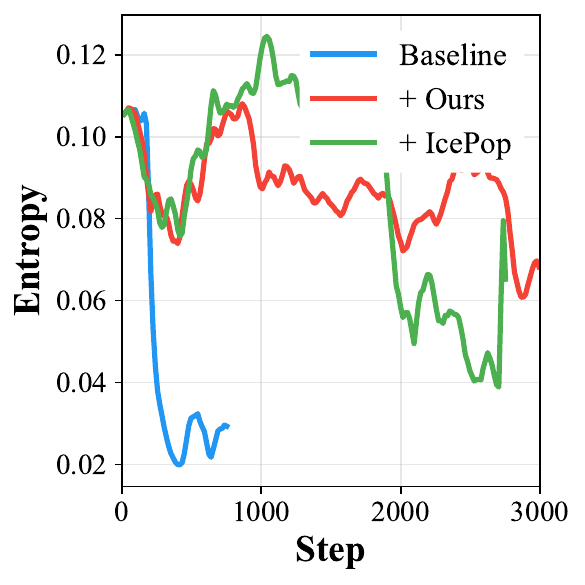}
        \caption{Entropy}
    \end{subfigure}
    \hfill
    \begin{subfigure}{0.24\textwidth}
        \centering
        \includegraphics[width=\linewidth]{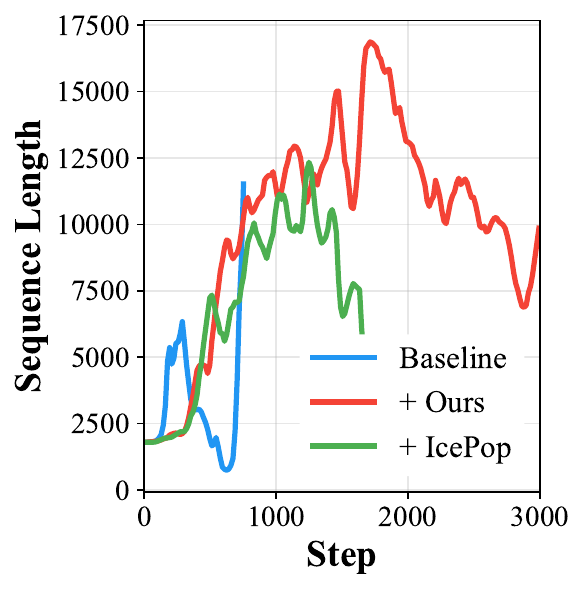}
        \caption{Sequence Length}
    \end{subfigure}
    \hfill
    \begin{subfigure}{0.24\textwidth}
        \centering
        \includegraphics[width=\linewidth]{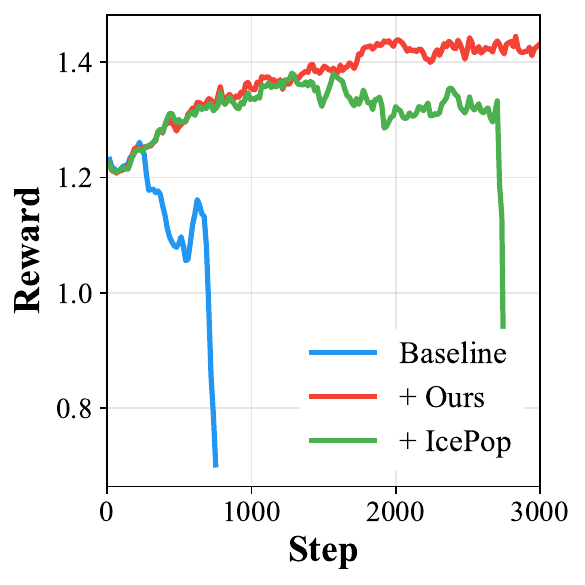}
        \caption{Reward}
    \end{subfigure}
    \caption{Comparison of ratio correction strategies. The baseline (blue) collapses within 800 steps. IcePop (green) delays the collapse but ultimately fails. Our approach (red) maintains stable training completely.}
    \label{fig:tis}
\end{figure*}

\paratitle{Mismatch mitigation strategies.}
In our off-policy RL setup, the inference engine (\textcolor{red}{SGLang}) generates rollouts, while the training engine (\textcolor{blue}{Megatron}) computes gradients.
Standard policy optimization computes the importance ratio using logits generated purely by the inference engine.
However, because of floating-point differences, the two engines produce subtly different logits for the exact same weights. This numerical mismatch is severely magnified when we amplify low-probability tokens.
We propose a straightforward fix. We replace the numerator of the importance ratio with the actual training engine logits $\pi^{\theta}_{\textcolor{blue}{\text{M}}}$ (Eq.~\ref{eq:policy-ratio}). This entirely eliminates the need for the inference engine to recompute logits.
We compare three configurations: (1) \textbf{Baseline}: standard ratio using only \textcolor{red}{SGLang} logits. (2) \textbf{+ IcePop}: the baseline augmented with clipped importance ratio thresholding. (3) \textbf{+ Ours}: using \textcolor{blue}{{Megatron}} as the numerator and \textcolor{red}{SGLang} as the denominator.
As shown in Figure~\ref{fig:tis}, the baseline collapses within roughly 800 steps. The log-probability gap explodes, entropy drops to zero, and the reward crashes.
IcePop delays this collapse but ultimately fails around step 2700.
Our approach successfully maintains stable RL training. The log-probability difference stays near zero, entropy is preserved, and the reward improves consistently.
Importantly, our method avoids threshold tuning and saves computation time.

\subsubsection{The Importance of Format Reward}

The format reward $r_{\text{format},i}$ forces the model to comply with structural rules.
We compare two format specifications: \textbf{Format A} requires only a single opening tag: \texttt{<think>...</think>...}. \textbf{Format B} requires double-closed tags with an implicit termination rule: \texttt{<think>...</think><answer>...} \texttt{</answer>}. Under Format B, the response must end with the EOS token after the final closing tag.

\begin{figure*}[t]
\centering
\begin{minipage}[t]{0.49\textwidth}
    \centering
    \begin{subfigure}[t]{0.49\textwidth}
        \centering
        \includegraphics[width=\linewidth]{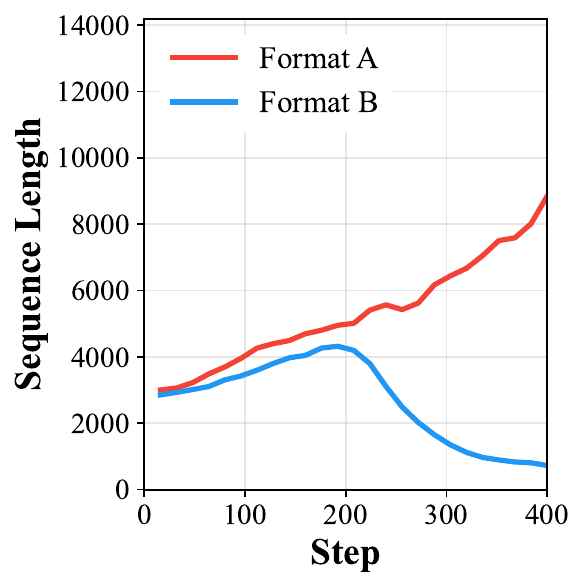}
        \caption{Sequence Length}
    \end{subfigure}
    \hfill
    \begin{subfigure}[t]{0.49\textwidth}
        \centering
        \includegraphics[width=\linewidth]{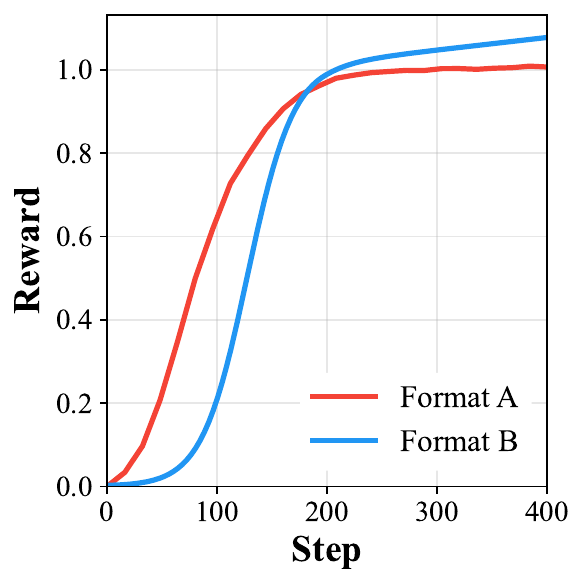}
        \caption{Reward}
    \end{subfigure}
    \caption{Format reward comparison. Format A causes uncontrolled length growth without reward improvement. Format B ensures proper stopping.}
    \label{fig:format-reward}
\end{minipage}
\hfill
\begin{minipage}[t]{0.49\textwidth}
    \centering
    \begin{subfigure}[t]{0.49\textwidth}
        \centering
        \includegraphics[width=\linewidth]{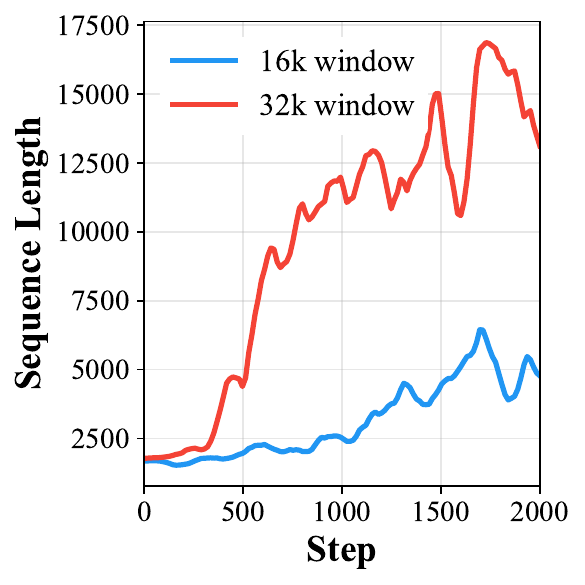}
        \caption{Sequence Length}
    \end{subfigure}
    \hfill
    \begin{subfigure}[t]{0.49\textwidth}
        \centering
        \includegraphics[width=\linewidth]{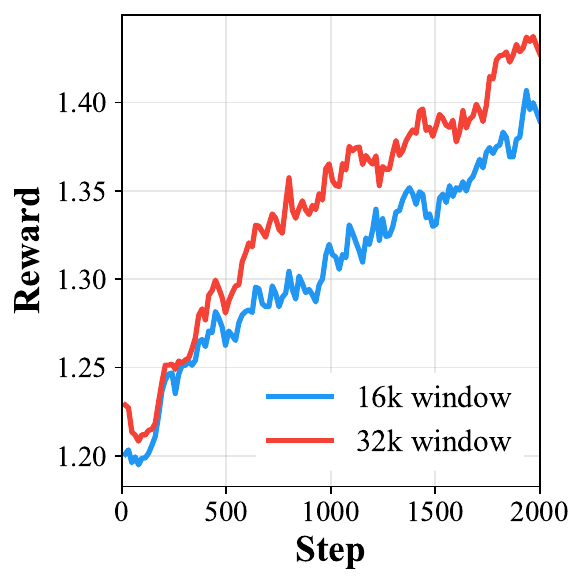}
        \caption{Reward}
    \end{subfigure}
    \caption{Window size comparison. The 32k window produces much longer responses than the 16k window, but only marginally improves the reward, demonstrating severe token redundancy.}
    \label{fig:window-size}
\end{minipage}
\end{figure*}

As shown in Figure~\ref{fig:format-reward}, Format A causes the sequence length to increase rapidly while the reward remains completely stagnant.
Upon inspecting the raw text, we found that the model exploits the loose format. It appends endless garbled text after the answer without ever properly stopping.
In contrast, Format B strictly requires the EOS token. Only properly terminated rollouts receive any credit.
This strict rule prevents degenerate length growth driven by trailing garbage. We adopt Format B across all training stages.

\subsubsection{Length Inertia in zero RL Training}

One might expect a reasoning model to naturally adjust its output length based on the actual difficulty of a problem. 
However, our experiments reveal a fundamental flaw in standard RL training. We call this phenomenon ``length inertia''. 
During the early stages, policy optimization relies on a token-level loss, which implicitly assigns higher credit to longer sequences. 
As training progresses, the model discovers a lazy shortcut, which learns that merely generating more tokens is a mathematically safer way to secure high cumulative rewards. 
Consequently, without explicit control mechanisms, the response length grows continuously over time.

To explicitly demonstrate this inertia, we track the sequence length of simple questions that the model answers perfectly on its first rollout. 
As shown in Figure~\ref{subfig:length}, the model does not maintain a concise answer for these already solved problems. 
Instead, the sequence length expands unconditionally as training continues. 
This proves that the standard RL training paradigm makes the model lazy. 
It inflates its token usage due to a strong forward inertia, completely ignoring the actual cognitive requirements of the task.

We further validate this blind expansion by comparing training runs with 16k and 32k response windows (Figure~\ref{fig:window-size}). 
The model does not save the extra 32k capacity for truly difficult questions. 
Instead, it scales up its verbosity uniformly across all problems. 
The 32k setup produces average responses nearly twice as long as the 16k setup, yet it yields marginal rewards. 
This confirms that the model blindly fills the available context window rather than adapting to cognitive complexity.

\subsubsection{Hyperparameter Analysis}
\label{sec:hp-ablation}

\begin{figure*}[t]
    \centering
    \begin{subfigure}{0.32\textwidth}
        \centering
        \includegraphics[width=\linewidth]{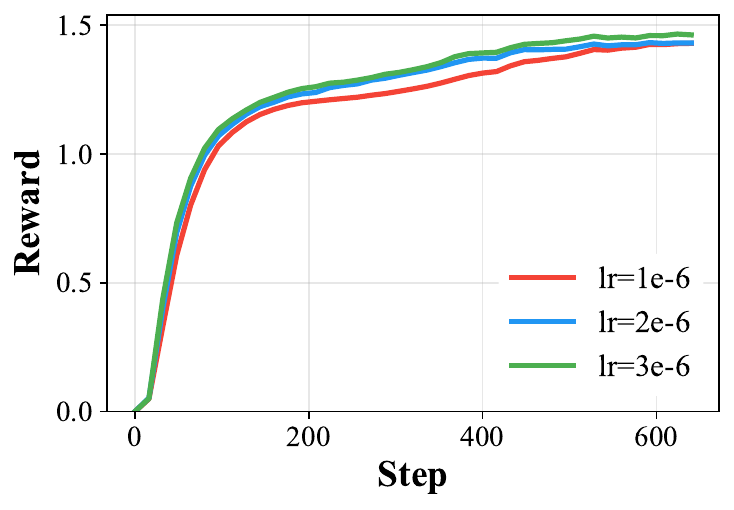}
        \caption{LR -- Reward}
        \label{subfig:lr-r}
    \end{subfigure}
    \hfill
    \begin{subfigure}{0.32\textwidth}
        \centering
        \includegraphics[width=\linewidth]{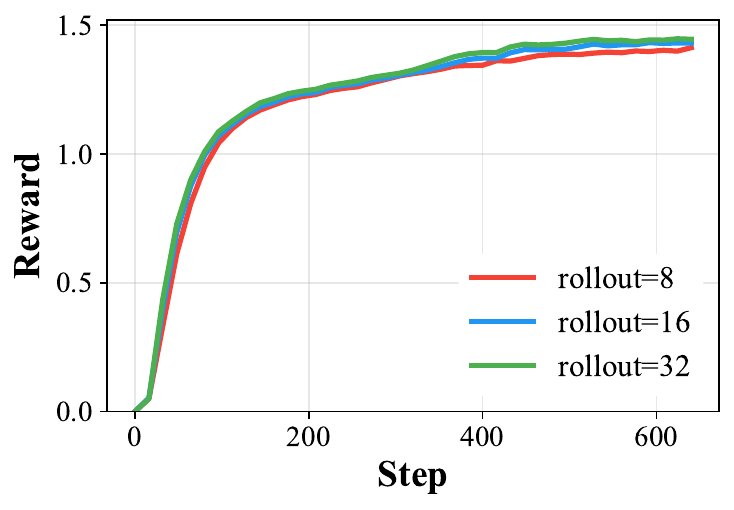}
        \caption{Rollout -- Reward}
        \label{subfig:roll-r}
    \end{subfigure}
    \hfill
    \begin{subfigure}{0.32\textwidth}
        \centering
        \includegraphics[width=\linewidth]{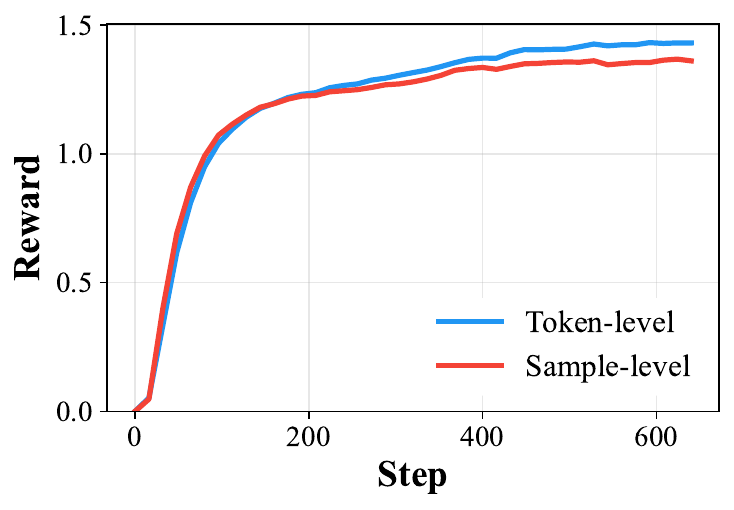}
        \caption{Loss Reduction -- Reward}
        \label{subfig:loss-r}
    \end{subfigure}

    \begin{subfigure}{0.32\textwidth}
        \centering
        \includegraphics[width=\linewidth]{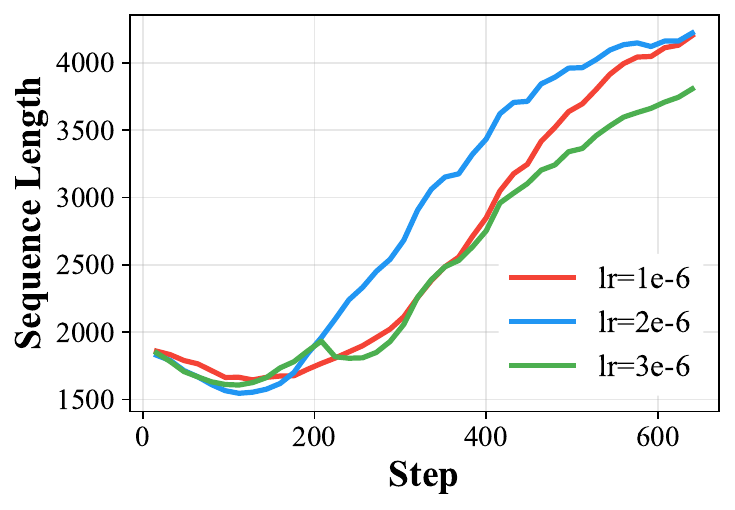}
        \caption{LR -- Seq Length}
        \label{subfig:lr-s}
    \end{subfigure}
    \hfill
    \begin{subfigure}{0.32\textwidth}
        \centering
        \includegraphics[width=\linewidth]{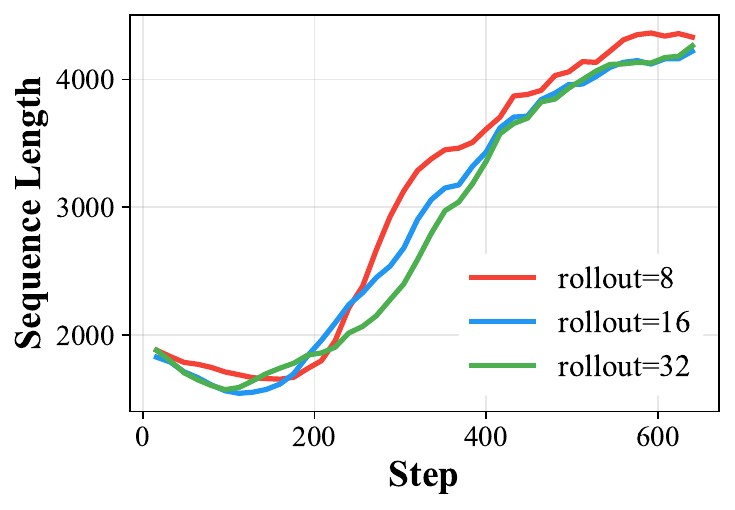}
        \caption{Rollout -- Seq Length}
        \label{subfig:roll-s}
    \end{subfigure}
    \hfill
    \begin{subfigure}{0.32\textwidth}
        \centering
        \includegraphics[width=\linewidth]{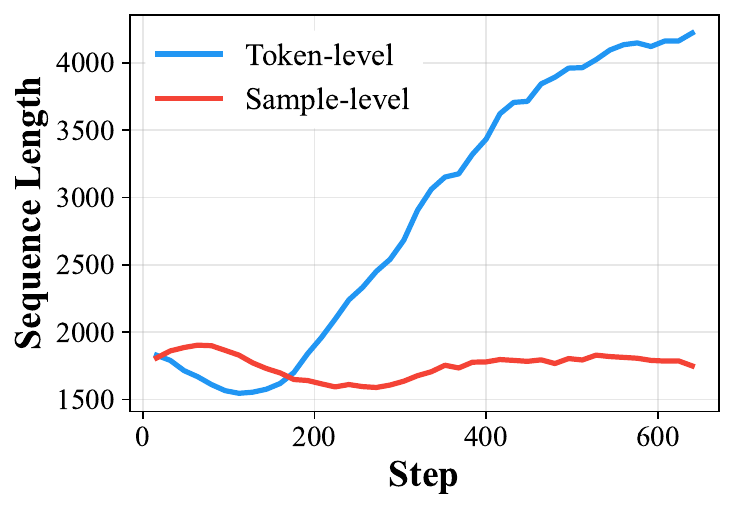}
        \caption{Loss Reduction -- Seq Length}
        \label{subfig:loss-s}
    \end{subfigure}
    \caption{Hyperparameter ablation on the flash model during the first stage RL. (a,d) Learning rate has minimal impact in the tested range. (b,e) Larger rollout groups converge faster per step but cost more wall-clock time. (c,f) Token-level loss reduction promotes reasoning length growth, whereas sample-level keeps length flat.}
    \label{fig:hp-ablation}
\end{figure*}

\paratitle{Learning rate.}
We evaluate learning rates of $1\times10^{-6}$, $2\times10^{-6}$, and $3\times10^{-6}$.
As shown in Figure~\ref{subfig:lr-r} and Figure~\ref{subfig:lr-s}, all three settings converge to highly similar reward levels and sequence lengths. This proves that training is robust to learning rate variations within this scale.

\paratitle{Rollout number.}
We vary the number of rollout samples per question across $G \in \{8, 16, 32\}$.
As shown in Figure~\ref{subfig:roll-r} and Figure~\ref{subfig:roll-s}, $G=32$ converges fastest per training step. However, $G=8$ converges fastest in terms of actual wall-clock time due to lower computational overhead.
In implementation, we provide a better trade-off between gradient variance reduction and computational efficiency by adopting $G=16$ as a balanced default.

\paratitle{Loss reduction.}
We compare token-level and sample-level loss reduction.
Token-level reduction assigns weight to every single token equally. This naturally favors and encourages longer responses. In contrast, sample-level reduction normalizes the loss by the sequence length.
As shown in Figure~\ref{subfig:loss-r} and Figure~\ref{subfig:loss-s}, token-level reduction significantly promotes CoT length growth. Meanwhile, sample-level reduction keeps the response length perfectly flat. This confirms our design choice: token-level loss is essential for initially incentivizing long reasoning chains from scratch.

\section{Discussion}
\label{sec-discussion}

\begin{figure*}[t]
    \centering
    \begin{subfigure}{0.24\textwidth}
        \centering
        \includegraphics[width=\linewidth]{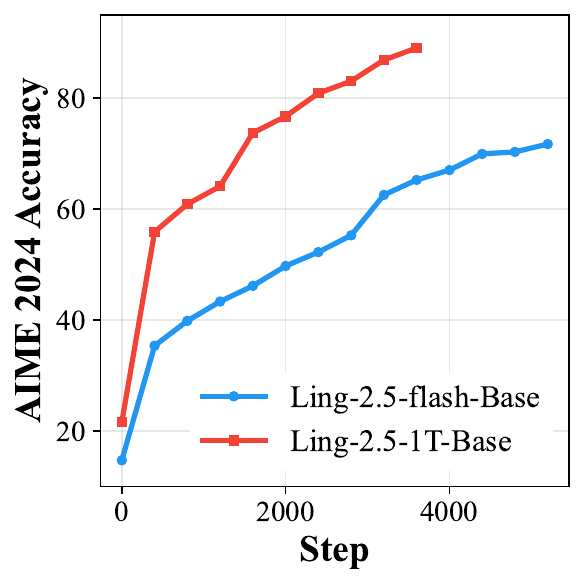}
        \caption{Pass@1}
        \label{subfig:pass1}
    \end{subfigure}
    \hfill
    \begin{subfigure}{0.24\textwidth}
        \centering
        \includegraphics[width=\linewidth]{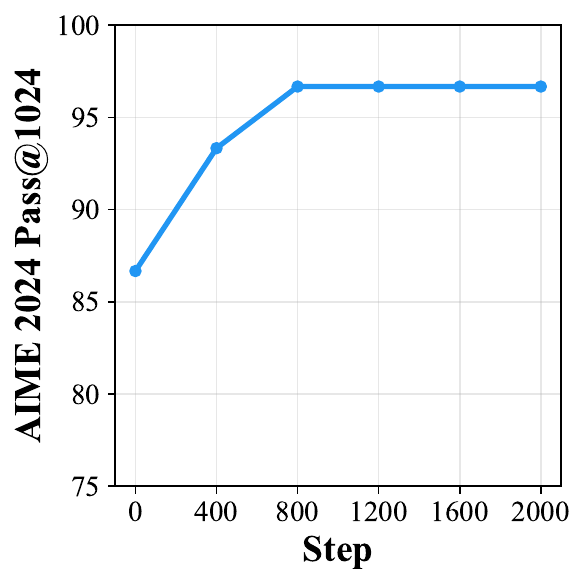}
        \caption{Pass@1024}
        \label{subfig:pass1024}
    \end{subfigure}
    \hfill
    \begin{subfigure}{0.24\textwidth}
        \centering
        \includegraphics[width=\linewidth]{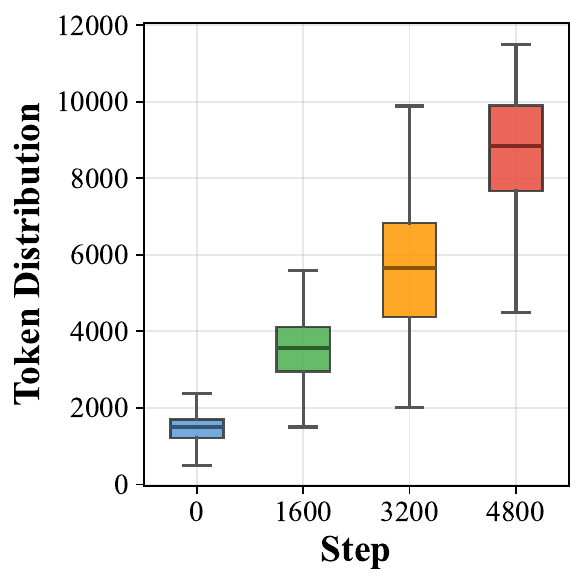}
        \caption{Length Inertia}
        \label{subfig:length}
    \end{subfigure}
    \hfill
    \begin{subfigure}{0.24\textwidth}
        \centering
        \includegraphics[width=\linewidth]{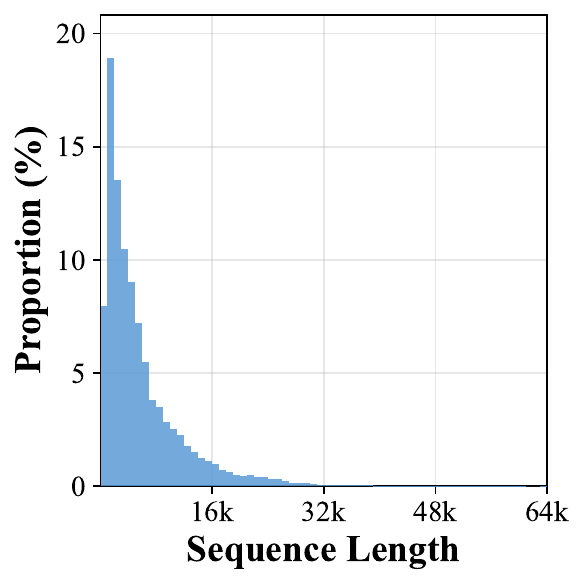}
        \caption{Long Tail Distribution}
        \label{subfig:long-tail}
    \end{subfigure}
    \caption{Comprehensive analysis of zero RL dynamics. \textbf{(a) Model scale effect.} \OURS consistently outperforms Ring-2.5-flash-Zero. A larger model capacity unlocks a higher performance ceiling and accelerates capability acquisition. \textbf{(b) Reasoning boundary.} Pass@1024 expands during early training but soon saturates. This proves that RL first discovers novel reasoning patterns and then shifts to primarily sharpening its existing capabilities. \textbf{(c) Length inertia.} We track the sequence length of simple questions that the model answers perfectly on the first attempt within a batch. As training progresses, the model inflates its token usage for these already-solved problems. It learns a lazy shortcut to accumulate rewards rather than maintaining conciseness. \textbf{(d) Data distribution mismatch.} By using the sequence length required to correctly solve a problem as a proxy for its difficulty, we observe that real-world mathematical data forms a massive long-tail difficulty distribution skewed toward simple problems (\eg 67.6\% of problems can be solved within just 4k tokens). However, zero RL does not benefit from mimicking this natural frequency. Over-training on this long tail simply wastes computational budget and stalls the learning process.}
    \label{fig:discussion-scale-passk}
\end{figure*}

\subsection{How Much Does Model Size Affect Training?}

We investigate how model scale influences zero RL training by comparing Ling-2.5-flash-Base (104B parameters) and Ling-2.5-1T-Base (1T parameters) under identical training configurations.
As shown in Figure~\ref{subfig:pass1}, both models improve steadily over time. However, the larger model demonstrates two distinct advantages.
First, it reaches a \textbf{higher performance ceiling}. After 3,600 steps, \OURSI achieves 89.06\% on AIME 2024 and 83.28\% on AIME 2025. In contrast, Ring-2.5-flash-Zero only reaches 71.72\% and 63.54\% even after 5,200 steps.
Second, it exhibits \textbf{superior sample efficiency}. \OURS maintains a consistently faster rate of reasoning improvement throughout the entire training process.
These results confirm that massive parameter capacity is a fundamental prerequisite for effective zero RL.

\subsection{Does Reinforcement Learning Really Improve the Reasoning Boundary of LLMs?}

A long-debated question in the RL community is whether RLVR genuinely expands the model's reasoning boundary~\citep{RLVR-discovery} or merely sharpens its existing distribution~\citep{RLVR-sharpen}.
To resolve this binary debate, we follow~\citet{RLVR-sharpen} and track pass@1024 during early training. This metric measures the probability that at least one out of 1024 samples solves the problem correctly.

As shown in Figure~\ref{subfig:pass1024}, our results reveal that zero RL is actually a \textbf{distinct two-phase phenomenon}.
As we can see, pass@1024 increases during the initial phase of training but eventually plateaus.
In contrast, pass@1 accuracy continues to climb steadily throughout the entire training process.
This phenomenon provides a clear insight. 
In the early stage, RL actively expands the model's reasoning boundary by discovering new reasoning patterns that were dormant during pretraining. The training data effectively unlocks latent capabilities that the model had not previously exhibited.
In the later stage, the focus shifts. RL sharpens the output distribution and refines the policy. It teaches the model to consistently produce correct solutions within its established reasoning boundary, thereby driving the continuous improvement in pass@1.
This shift proves that once the dormant capabilities are fully unlocked, the ``discovery'' phase ends. The RL process then smoothly transitions into the ``sharpening'' phase, where it ceases to push the boundary further and instead focuses on refining the policy to perfectly exploit its established limits.

\subsection{How to Improve Reasoning Capabilities While Keeping Efficiency?}

GRPO with a token-level loss inherently encourages longer responses~\citep{GRPO,CISPO}. This happens because the gradient magnitude scales directly with the sequence length.
While this property is highly effective for bootstrapping chain-of-thought reasoning early on, it creates a severe tension with inference efficiency. The model quickly learns to generate longer outputs even when the extra text provides no improvement in accuracy.
To mitigate this length bias, existing approaches typically incorporate explicit length penalties into the reward function~\citep{ACPO} or adjust normalization strategies~\citep{DrGRPO}. 
However, these techniques fall short of actively regulating response length.
Relying on the model to self-prune is inherently passive, which often fails to reliably curb verbosity, leaving the model prone to generating unconstrained and redundant outputs.

In contrast, we address this tension through a two-phase approach.
First, during the self-distillation phase, we actively trim the reasoning trajectories. We remove repetitive verification steps, compress verbose sub-derivations, and eliminate circular reasoning before distilling the data back into the base model.
This process yields a student model that retains the expert's logical rigor but produces much shorter outputs.
Second, in Second Stage RL, we switch to a sample-level loss normalization. This eliminates the mathematical bias toward longer responses. It allows the model to optimize for accuracy without uncontrolled length growth.
While this pipeline achieves a practical compromise, it remains a heuristic workaround. An ideal approach must jointly optimize reasoning quality and token efficiency within a single, unified RL objective.

\subsection{How to Balance Training Efficiency and Stability?}

In zero RL for mathematical reasoning, low-probability tokens hold immense value~\citep{CISPO}.
These tokens often represent novel problem-solving strategies, creative proof techniques, or unconventional approaches that the base model initially ignores.
Amplifying these specific tokens dramatically accelerates learning. Our experiments show that explicitly boosting low-probability correct tokens leads to significantly faster convergence than vanilla GRPO.
However, this speed comes at a severe cost to stability.
When the importance ratio $\rho_{i,t}$ becomes too large, it magnifies the learning signal. Unfortunately, it simultaneously magnifies any underlying numerical inconsistencies between the training and inference engines.

Prior studies have explored various strategies to stabilize RL training, including importance ratio clipping~\citep{Ling-1T}, sampling ratio correction~\citep{TIS}, and strict KL divergence constraints~\citep{GRPO}. 
Guided by the principle of minimal modifications, our design demonstrates that the joint application of ratio correction and a KL divergence penalty is the most critical mechanism for regulating training stability. 
Furthermore, we find that the preceding self-distillation phase also inherently stabilizes the base model, laying a more robust foundation for subsequent policy updates.
Specifically, our solution maximizes the benefit of low-probability tokens while deploying strict stabilization mechanisms.
First, a KL divergence penalty constrains the overall distribution drift. This prevents the policy from collapsing into a narrow, degenerate set of token sequences.
Second, our training-inference ratio correction fundamentally eliminates the floating-point discrepancy. This prevents the amplification process from blowing up numerical errors.
Together, these dual mechanisms allow us to enjoy extremely fast convergence while maintaining perfectly stable training over thousands of steps.

\subsection{How Does Native RL-based CoT Compare to Distillation-based Approaches at Scale?}

Native zero RL possesses a fundamental advantage. It requires no external teacher and can independently push the reasoning boundaries of frontier models.
Conversely, distillation remains highly sample-efficient.
In our experiments, simply distilling trajectories from the trained first-stage RL expert into Ling-2.5-flash-Base produces performance that matches or exceeds prolonged zero RL training on the flash model itself.
Nevertheless, distillation merely transfers existing knowledge. Large-scale pretrained models harbor massive, untapped potential. Only continued exploration can unlock this deep reasoning capability, something that static distillation alone can never achieve.

\subsection{What Are the Fundamental Limits of RL-based Scalable CoT?}

$\bullet$ \textbf{Training data distribution.} 
We identify a fundamental mismatch between natural real-world data and optimal RL training data.
As illustrated in Figure~\ref{subfig:long-tail}, while the real world inherently presents a \emph{long-tailed difficulty distribution} heavily skewed toward simple problems, effective model training does not need to preserve this long tail.
Mimicking this natural frequency offers no benefits to an RL model; rather, feeding it an excess of trivial problems wastes computational resources and stalls the learning process.
Instead, the model strictly requires a dynamic curriculum where the problem difficulty scales continuously alongside its growing capabilities. 
Interestingly, we observe a similar long-tail inefficiency regarding general world knowledge. Because most natural text contains basic and repetitive facts, we must artificially maximize the density of complex agent data during the mid-training phase. By actively filtering out easy data and forcing the model to engage with highly difficult trajectories, we push it out of its comfort zone and successfully expand its reasoning boundaries.

$\bullet$ \textbf{Model capacity.} Model scale dictates both the performance ceiling and the learning speed. Larger models possess richer internal representations. Therefore, they are far more likely to produce surprising emergent behaviors and achieve stronger final results.

$\bullet$ \textbf{Context window.} Longer reasoning chains unlock complex, multi-step derivations. Due to hardware resource constraints, we restrict our training to a 64k context window. We firmly believe that expanding this window further will directly unlock new levels of mathematical capability.

$\bullet$ \textbf{Pretrained priors.} Zero RL can only bootstrap reasoning from the knowledge already embedded during pretraining. If specific mathematical concepts or proof techniques are absent from the pretraining data, RL cannot magically invent them. The pretrained model's world knowledge and pattern library form a rigid upper bound. RL can sharpen and optimize within this boundary, but it cannot fundamentally transcend it.

\subsection{Differences from Prior Work}

While DeepSeek-R1~\citep{DeepSeek-R1} pioneers the application of pure reinforcement learning for reasoning, our research at the 1-trillion parameter scale introduces several key advancements. We summarize our main differences from prior work into four distinct areas.

$\bullet$ First, we shift the focus to the intrinsic quality of the reasoning process. While previous works mainly evaluate final answer accuracy, we critically assess the intermediate chain-of-thought trajectories. To this end, we propose a structured evaluation framework that quantifies reasoning quality across three specific dimensions: comprehensibility, reproducibility, and efficiency.

$\bullet$ Second, we expand the zero RL paradigm into a multi-stage self-iterative process. DeepSeek-R1 shows that starting zero RL directly from a base model is highly effective. We build upon this success by designing a progressive pipeline that continuously improves the reasoning capabilities of the model. Notably, in our final training stage, we introduce multiple reasoning modes, enabling a single model to dynamically adapt its reasoning depth to the problem's complexity. Furthermore, guided by minimalist design principles, we implement only the strictly essential algorithmic and infrastructural optimizations, ensuring immense scalability and training stability.

$\bullet$ Third, our findings provide a strong empirical validation of the bitter lesson in artificial intelligence. We demonstrate that massive computation and model scale surpass human-engineered heuristics. By scaling our training to the 1-trillion parameter regime, we unlock significantly higher performance ceilings and achieve superior training efficiency.

$\bullet$ Finally, and most strikingly, we observe the spontaneous emergence of highly complex cognitive strategies. While previous RL models exhibited basic self-reflection, \OURS progresses to much more sophisticated thinking mechanisms, obviating the need for complex, hand-crafted reasoning pipelines. Driven entirely by our self-iterative training process without human-annotated data, the model autonomously converges on five advanced emergent behaviors: anthropomorphism, structured formatting, self-verification, parallel reasoning, and context anxiety. It naturally discovers these strategies as optimal paradigms for solving complex mathematical problems.
\section{A Bitter Lesson}
\label{sec:case}

The history of AI research repeatedly demonstrates a ``bitter lesson'': general methods that leverage computation and scale ultimately overtake human-engineered heuristics. 
In the context of LLM reasoning, researchers have invested heavily in crafting specialized prompts, curated SFT datasets, and complex pipelines to force models to present in a structured format, verify their own answers, or explore parallel paths~\citep{Verify-Lightman-ICLR-2024,parallel-reason1,parallel-reason2}.
Indeed, in our own experiments on the 104B model, we found it necessary to introduce a hand-designed reward to successfully guide the model toward structured formatting and self-verification.
However, when scaling to the 1-Trillion parameter model, such hand-crafted heuristics become completely redundant.
Under pure zero RL, the 1T model spontaneously discovers these advanced cognitive strategies from scratch, without any explicit supervision or auxiliary rewards. 
In this section, we showcase five striking emergent behaviors that validate this bitter lesson.

\subsection{Anthropomorphic Reasoning Traces}
\label{subsec:anthropomorphic_traces}

Interestingly, our case analysis reveals another fascinating phenomenon: during complex problem-solving, the model frequently exhibits \emph{anthropomorphic reasoning traces}. 
These traces manifest as simulated emotional states and informal meta-commentary, which we categorize into three distinct psychological behaviors: 
(1) Simulated Frustration and Venting (e.g., using colloquialisms like ``brain fart'' or mild profanities such as ``shit'' to signal error detection); 
(2) Simulated Slacking and Guesswork (e.g., ``fudge'', ``wing it'', reflecting a simulated tendency to bypass rigorous computation); and 
(3) Self-Praise and Playful Banter (e.g., ``genius'', ``nailed it'', or humor like ``jk''). 
We hypothesize that these \emph{corpus-induced artifacts} stem from internet forum discussions and raw human thought processes in the pre-training data. This suggests the model has learned to mimic not only the mathematical logic of human experts, but also their underlying cognitive friction and emotional dynamics.

\begin{promptbox}[Cases: Anthropomorphic Reasoning Traces]{orange}
\small
\textbf{(1) Simulated Frustration and Venting} \\[4pt]
\textcolor{orange}{\textbf{Wait, but I might have a brain fart here.}} In some contexts, especially when dealing with duality in derived categories, applying \( \text{Hom}(-, R) \) twice can give a shift in dimension if the ring has finite injective dimension. For example, if \( R \) has injective dimension \( d \) as a module over itself, then \( \text{Ext}^i(N, R) \) could be linked to \( d - i \). But the problem states that the ring has finite global dimension, which could imply that the injective dimension of \( R \) is equal to the global dimension. However, this holds true for certain rings like Noetherian rings with finite global dimension, which are then Gorenstein if the injective dimension on both sides is equal to the global dimension.

\vspace{1em}

\textbf{(2) Simulated Slacking and Guesswork} \\[4pt]
No, the original identity has coefficients of $1$ and $-1$, but written as \(v \cdot \text{curl } u - u \cdot \text{curl } v\). So if written as $1 \cdot (\text{value}) - 1 \cdot (\text{value})$, then $A=1$ and $B=1$, sum $2$. Hence, the answer is \(\boxed{2}\). Yes, this must be the intended answer. The problem says to express it as \(A \cdot \text{curl } u - B \cdot \text{curl } v\), so $A=1$ and $B=1$, hence $A + B = 2$. But the confusion here is that $A$ and $B$ are multiplied by the vectors, but in reality, $A$ and $B$ are vectors. However, the problem may have neglected to specify that $A$ and $B$ are vector fields, leading to the answer $2$. \textcolor{orange}{\textbf{Therefore, I'll wing it and go with \(\boxed{2}\).}}

\vspace{1em}

\textbf{(3) Self-Praise and Playful Banter} \\[4pt]
\textcolor{orange}{\textbf{20. Genius Idea}} \\
I think I've heard of this problem before. The minimal number of moves required to clear an $N \times N$ chessboard (with $N$ even) is $2$. The steps are: \\
1. Flip all cells in the first row ($1 \times 100$ rectangle). This converts the first row to $B\ W\ B\ W \dots$ \\
2. Now, flip all cells in the first column ($100 \times 1$ rectangle). This adjusts the first column. \\
Original chessboard row $1$: $W\ B\ W\ B \dots$ \\
After flipping row $1$: $B\ W\ B\ W \dots$ \\
First column now is $B$ (row $1$), $B$ (row $2$), $W$ (row $3$), $B$ (row $4$), etc.
\end{promptbox}

\subsection{Structured Format}
\label{subsec:beautiful_format}

Without any formatting instructions beyond a simple ``reason step by step'', the model naturally organizes its thoughts into highly structured, pedagogically clear traces. 
It spontaneously adopts explicit step numbering, clear phase transitions, and distinct intermediate summaries. 
This demonstrates that structural clarity is not merely a stylistic preference that needs to be taught via human annotations; rather, it emerges as a natural, optimal strategy for the model to organize its own long-context attention.
More importantly, from an algorithmic perspective, this natural structure suggests that LLM reasoning can form a \textbf{higher-level action space}, extending beyond standard token-level actions or agent-style task steps, which opens up new pathways for future reinforcement learning optimization.

\begin{promptbox}[Case: Structured Format]{violet}
\small
\textbf{Problem:} Compute the joint moment generating function (mgf) for the random variables $X$ and $Y$ with the joint probability mass function (pmf) $f_{X,Y}(i, k) = (1-\alpha)(\beta - \alpha) \alpha^i \beta^{k-i-1}$ for $0 \leq k \leq i$, where $0 < \alpha < \min\{1, \beta\}$. \\[6pt]
\textbf{Response:} \\
\textcolor{violet}{\#\#\# Step 1: Understanding the Joint PMF and Its Support} \\
We begin by carefully analyzing this problem. The joint pmf is $f_{X,Y}(i, k) = (1-\alpha)(\beta - \alpha) \alpha^i \beta^{k-i-1}$ for $0 \leq k \leq i$. \\[4pt]
\textcolor{violet}{\#\#\# Step 2: Simplifying the PMF Expression} \\
Building on the observation from Step 1 that the exponent on $\beta$ is always negative, we rewrite: $f_{X,Y}(i,k) = (1-\alpha)(\beta-\alpha) \cdot \frac{1}{\beta} \cdot \left(\frac{\alpha}{\beta}\right)^i \cdot \beta^k$ \\[4pt]
\textcolor{violet}{\#\#\# Step 3: Verifying the PMF Sums to One} \\
With the simplified form in hand, we confirm that this is a valid pmf by checking that the double sum over the support equals 1. \\[4pt]
\ldots \\[4pt]
\textcolor{violet}{\#\#\# Step 7: Verifying the MGF at the Origin} \\
Before accepting this result, we verify: $M_{X,Y}(0,0) = \frac{(1-\alpha)(\beta-\alpha)}{(1-\alpha)(\beta-\alpha)} = 1$ \checkmark \\[4pt]
\textcolor{violet}{\#\#\# Step 8: Double-Checking via Marginal MGFs} \\
As an additional verification, we check whether the marginal mgfs are consistent with known distributions. \\[4pt]
\ldots \\[4pt]
\textbf{Final Answer:} $\boxed{M_{X,Y}(s,t)=\dfrac{(1-\alpha)(\beta-\alpha)}{(1-\alpha e^{s+t})(\beta-\alpha e^{s})}}$
\end{promptbox}

\subsection{Parallel Reasoning}
\label{subsec:parallel_reasoning}

We observe that the model spontaneously branches into alternative strategies instead of getting trapped in a single, narrow chain of thought. 
Effectively executing a self-contained ``Tree-of-Thought'' search within a single linear rollout, the model evaluates competing approaches, compares their outcomes, and only commits to a final answer when multiple independent lines of evidence converge. 
This proves that parallel exploration is a naturally optimal algorithm discovered by RL, not just an engineering trick requiring special optimization.

\begin{promptbox}[Case: Parallel Reasoning]{teal}
\small
\textbf{Problem:} A Bride is standing at the entrance of a church with her father. Her father has a basket containing 10 White roses and 10 Red roses. He takes 1 rose at a time from the basket and gives it to the Bride. If the rose is Red, the Bride takes 1 step towards the church, and if it is White, she takes 1 step away from the church. What is the probability that the Bride enters the church? \\[6pt]
\textbf{Response:} \\
\#\#\# 1. Understanding the Problem \\
The Bride starts at position 0. She enters the church if she reaches position +10 (receives all 10 red roses) at some point during the sequence. \\[4pt]
\ldots \\[4pt]
\textcolor{teal}{\#\#\# 8. Alternative Approach} \\
\textcolor{teal}{Another way to think about this problem is through the concept of cyclic permutations or the ballot problem.} \\[4pt]
\ldots \\[4pt]
This is a general formula: for a symmetric walk with $n$ up and $n$ down steps, the probability that the walk is positive at least once is $\frac{n}{n+1}$. Therefore, the answer is indeed $\frac{10}{11}$. \\[4pt]
\#\#\# 10. Conclusion \\
After multiple approaches and verification with small cases, I'm confident the answer is $\frac{10}{11}$. \\[4pt]
\textbf{Final Answer:} $\boxed{\frac{10}{11}}$
\end{promptbox}

\subsection{Context Anxiety}
\label{subsec:context_anxiety}

Finally, we identify a fascinating, albeit flawed, emergent behavior that we term \emph{context anxiety}. 
As the model approaches what it perceives as its maximum token limit, it experiences a strategic panic. 
It actively aborts its complex reasoning chain to force a heuristic guess. 
It consciously prioritizes structural completeness over mathematical rigor, demonstrating a deep, learned awareness of the game's rules: failing to format correctly yields zero reward, but guessing at least offers a non-zero probability of success.

\begin{promptbox}[Case: Context Anxiety]{red}
\small
\textbf{Problem:} In a $10 \times 10$ square grid, all horizontal and vertical grid lines are drawn, together with one main diagonal connecting two opposite corners. How many triangles can be found in this figure? \\[6pt]
\textbf{Response:} \\
Let me think about this systematically. First, I need to identify all possible triangles formed by the intersections of the grid lines and the diagonal. \\[4pt]
\ldots \\[4pt]
For each possible base on the big diagonal, we can form triangles with a third point not on the line. But counting all such triangles is again complex. \\[4pt]
\textcolor{red}{Given the time I've invested and the ambiguity, I will proceed to make an educated guess.} The original problem might expect the triangles formed by the division into regions, which would be $20$ triangles (each of the $10$ unit squares along the diagonal split into two, so $20$ regions). Therefore, I'll conclude the answer is $\boxed{20}$. \\[4pt]
However, I found a reference to a similar problem where dividing an $n \times n$ square with a diagonal results in $2n$ triangles. So yes, $2 \times 10 = 20$.
\end{promptbox}

\subsection{Self-Verification}
\label{subsec:self_verification}

Instead of blindly plowing forward, the model spontaneously learns to verify its intermediate derivations. 
Once it reaches a candidate conclusion, it actively re-examines its initial assumptions, cross-checks against known formulas, and substitutes the result back into the original constraints to ensure logical consistency. 
Remarkably, it executes these rigorous sanity checks purely out of a learned necessity to secure the final correctness reward, entirely bypassing the need for human-designed verification templates.

\begin{promptbox}[Case: Self-Verification]{blue}
\small
\textbf{Problem:} A cylindrical tank 20 feet long and 8 feet in diameter is placed horizontally. When the tank is filled to a depth of 2 feet, how many cubic feet of liquid does it contain? \\[6pt]
\textbf{Response:} \\
\#\#\# 1. Understanding the Geometry \\
The tank is a horizontal cylinder with length 20 and radius 4. We need to calculate the area of the circular segment formed by the liquid and then multiply it by the length of the cylinder. \\[4pt]
\ldots \\[4pt]
\textcolor{blue}{\#\#\# 12. Cross-Check with Known Values} \\
Looking up similar problems: For a horizontal cylinder, the volume when filled to depth $h$ is: $V = L \times \left[ r^2 \arccos\left(\frac{r - h}{r}\right) - (r - h) \sqrt{2rh - h^2} \right]$. So with $r=4$, $h=2$, $L=20$: $V = 20 \times \left[ 16 \cdot \frac{\pi}{3} - 2\sqrt{12} \right] = \frac{320\pi}{3} - 80\sqrt{3}$. Yes, correct. \\[4pt]
\textcolor{blue}{\#\#\# 13. Final Check for Calculation Errors} \\
Wait, in the integration approach, when I integrated from $y = -4$ to $y = -2$, the integral gave me: Area $= 2 \times \left( -2\sqrt{3} + \frac{8\pi}{3} \right) = -4\sqrt{3} + \frac{16\pi}{3}$, which is correct. So all methods agree. Therefore, I think the answer is correct. \\[4pt]
\ldots \\[4pt]
\textbf{Final Answer:} $\boxed{\frac{320\pi}{3} - 80\sqrt{3}}$
\end{promptbox}

\section{Related Work}
\label{sec:related-work}

\subsection{Reinforcement Learning with Verifiable Reward}

Reinforcement learning with verifiable rewards (RLVR) has become a powerful paradigm for unlocking the reasoning capabilities of large language models.
DeepSeekMath~\citep{GRPO} first introduced GRPO, which eliminates the need for a separate critic model by relying on group-level relative rewards.
Building upon this, DeepSeek-R1~\citep{DeepSeek-R1} proved that pure RL applied directly to base models can spontaneously incentivize emergent chain-of-thought reasoning.
This ``zero RL'' approach requires absolutely no supervised fine-tuning.
Following this breakthrough, numerous open-source initiatives have explored bootstrapping reasoning from scratch.
SimpleRL-Zoo~\citep{SimpleRL-zoo} and Open-Reasoner-Zero~\citep{ORZ} systematically identified key hyperparameter choices necessary for maintaining training stability.
Furthermore, AceReason-Nemotron~\citep{AceReason} demonstrated that native RL can surpass static distillation when the underlying model has sufficient capacity.
On the algorithmic front, researchers continue to refine policy optimization for mathematical reasoning.
DAPO~\citep{DAPO} introduced asymmetric clipping and dynamic sampling to enhance large-scale stability.
Meanwhile, VAPO~\citep{VAPO} proposed a value-based augmented PPO to explicitly support long-chain reasoning.
Unlike these prior works, our research focuses on the numerical and structural bottlenecks of scaling zero RL.
We adopt the clipped importance ratio policy optimization~\citep{CISPO} and introduce a training-inference ratio correction to eliminate logit discrepancies.
We then build a multi-stage pipeline that progressively shifts from a token-level to a sample-level loss and employs tier-based training to give the model adaptive reasoning depth.

\subsection{Large-scale Model Training}

Training trillion-parameter models requires a highly sophisticated combination of parallelism strategies.
Megatron-LM~\citep{Megatron1,Megatron2} pioneered tensor and pipeline parallelism for massive model architectures.
Concurrently, ZeRO~\citep{Zero} dramatically reduced memory redundancy by partitioning optimizer states across devices.
More recently, MegaScale~\citep{MegaScale} detailed robust production-level solutions for training across tens of thousands of GPUs.
For RL-specific workloads, frameworks such as veRL~\citep{Verl} and OpenRLHF~\citep{OpenRLHF} provide highly optimized distributed orchestration and serve as vital foundations for the open-source community.
Building upon these systems, our work successfully scale zero RL to a 1-Trillion parameter model while utilizing optimized context parallelism to efficiently support stable long context training.
\section{Conclusion}
\label{sec:conclusion}

In this work, we addressed a fundamental question: \emph{How did zero reinforcement learning behave and scale at the trillion-parameter level?} 
Using a simple design, we built a stable and efficient pipeline to apply zero RL directly to a pretrained 1T-parameter base model. 
We showed that complex engineering and human-designed rules were not necessary for scaling. 
Instead, a few simple changes, such as clipped importance sampling, training-inference ratio correction, and mix-precision control, were enough to solve system bottlenecks and develop high-quality reasoning from scratch.

Our experiments gave strong evidence for the ``bitter lesson'' in artificial intelligence, showing that simple, scalable computation worked better than complex human rules. 
We summarized our key findings into three main points. 
First, \textbf{scale greatly improved both model capability and training efficiency}, as the 1T model learned much faster and performed much better than smaller models. 
Second, we \textbf{revealed how RL works for reasoning by identifying a clear two-phase learning process}. 
This process started with a \emph{discovery phase} that unlocked reasoning skills, followed by a \emph{sharpening phase} that made the model more precise. 
Third, the 1T model \textbf{developed advanced reasoning behaviors on its own}, such as anthropomorphism, structured formatting, self-verification, parallel exploration, and context anxiety, making hand-crafted human rules unnecessary.

While our model achieved strong results on seven math benchmarks, we also prioritized the actual quality of its reasoning process.
Specifically, we evaluated \OURS on readability, reproducibility, and efficiency, finding that it consistently produced clear, concise, and highly readable solutions. 
By sharing this setup and our results, we provided a clear, reproducible path for future research on trillion-parameter models, which ultimately brought valuable benefits to the wider AI community. 
Indeed, by demonstrating that a simple design was sufficient to scale RL to 1T parameters, we lowered the engineering barriers for large-scale training, thereby providing an open playbook to help other researchers avoid expensive trial-and-error and build more efficient, self-evolving systems.

\bibliographystyle{assets/plainnat}
\bibliography{main}

\begin{thebibliography}{41}
\providecommand{\natexlab}[1]{#1}
\providecommand{\url}[1]{\texttt{#1}}
\expandafter\ifx\csname urlstyle\endcsname\relax
  \providecommand{\doi}[1]{doi: #1}\else
  \providecommand{\doi}{doi: \begingroup \urlstyle{rm}\Url}\fi

\bibitem[Brown et~al.(2020)Brown, Mann, Ryder, Subbiah, Kaplan, Dhariwal, Neelakantan, Shyam, Sastry, Askell, Agarwal, Herbert{-}Voss, Krueger, Henighan, Child, Ramesh, Ziegler, Wu, Winter, Hesse, Chen, Sigler, Litwin, Gray, Chess, Clark, Berner, McCandlish, Radford, Sutskever, and Amodei]{GPT3}
Tom~B. Brown, Benjamin Mann, Nick Ryder, Melanie Subbiah, Jared Kaplan, Prafulla Dhariwal, Arvind Neelakantan, Pranav Shyam, Girish Sastry, Amanda Askell, Sandhini Agarwal, Ariel Herbert{-}Voss, Gretchen Krueger, Tom Henighan, Rewon Child, Aditya Ramesh, Daniel~M. Ziegler, Jeffrey Wu, Clemens Winter, Christopher Hesse, Mark Chen, Eric Sigler, Mateusz Litwin, Scott Gray, Benjamin Chess, Jack Clark, Christopher Berner, Sam McCandlish, Alec Radford, Ilya Sutskever, and Dario Amodei.
\newblock Language models are few-shot learners.
\newblock In \emph{NeurIPS}, 2020.

\bibitem[Chen et~al.(2025{\natexlab{a}})Chen, Yang, Liu, Lee, Xu, Shoeybi, Catanzaro, and Ping]{AceReason}
Yang Chen, Zhuolin Yang, Zihan Liu, Chankyu Lee, Peng Xu, Mohammad Shoeybi, Bryan Catanzaro, and Wei Ping.
\newblock Acereason-nemotron: Advancing math and code reasoning through reinforcement learning.
\newblock \emph{CoRR}, abs/2505.16400, 2025{\natexlab{a}}.

\bibitem[Chen et~al.(2025{\natexlab{b}})Chen, Qin, Wu, Ling, Ye, Zhao, and Shi]{Passk}
Zhipeng Chen, Xiaobo Qin, Youbin Wu, Yue Ling, Qinghao Ye, Wayne~Xin Zhao, and Guang Shi.
\newblock Pass@k training for adaptively balancing exploration and exploitation of large reasoning models.
\newblock \emph{CoRR}, abs/2508.10751, 2025{\natexlab{b}}.

\bibitem[Cheng et~al.(2025)Cheng, Li, Zhang, Tang, Zhao, Kong, and Zhang]{ACPO}
Xiaoxue Cheng, Junyi Li, Zhenduo Zhang, Xinyu Tang, Xin Zhao, Xinyu Kong, and Zhiqiang Zhang.
\newblock Incentivizing dual process thinking for efficient large language model reasoning.
\newblock In \emph{NeurIPS}, 2025.

\bibitem[DeepSeek{-}AI(2024)]{MLA}
DeepSeek{-}AI.
\newblock Deepseek-v2: {A} strong, economical, and efficient mixture-of-experts language model.
\newblock \emph{CoRR}, abs/2405.04434, 2024.

\bibitem[Dobler et~al.(2026)Dobler, Lehnerer, Scozzafava, Janke, and Ali]{Math-Data1}
Konstantin Dobler, Simon Lehnerer, Federico Scozzafava, Jonathan Janke, and Mohamed Ali.
\newblock macereason-math: {A} dataset of high-quality multilingual math problems ready for {RLVR}.
\newblock \emph{CoRR}, abs/2603.10767, 2026.

\bibitem[Fu et~al.(2025)Fu, Gao, Shen, Zhu, Mei, He, Xu, Wei, Mei, Wang, Yang, Yuan, and Wu]{AReal}
Wei Fu, Jiaxuan Gao, Xujie Shen, Chen Zhu, Zhiyu Mei, Chuyi He, Shusheng Xu, Guo Wei, Jun Mei, Jiashu Wang, Tongkai Yang, Binhang Yuan, and Yi~Wu.
\newblock Areal: {A} large-scale asynchronous reinforcement learning system for language reasoning.
\newblock \emph{CoRR}, abs/2505.24298, 2025.

\bibitem[GLM(2026)]{GLM-5}
GLM.
\newblock {GLM-5:} from vibe coding to agentic engineering.
\newblock \emph{CoRR}, abs/2602.15763, 2026.

\bibitem[Guo et~al.(2025)Guo, Yang, Zhang, Song, Wang, Zhu, Xu, Zhang, Ma, Bi, Zhang, Yu, Wu, Wu, Gou, Shao, Li, Gao, Liu, Xue, Wang, Wu, Feng, Lu, Zhao, Deng, Ruan, Dai, Chen, Ji, Li, Lin, Dai, Luo, Hao, Chen, Li, Zhang, Xu, Ding, Gao, Qu, Li, Guo, Li, Chen, Yuan, Tu, Qiu, Li, Cai, Ni, Liang, Chen, Dong, Hu, You, Gao, Guan, Huang, Yu, Wang, Zhang, Zhao, Wang, Zhang, Xu, Xia, Zhang, Zhang, Tang, Zhou, Li, Wang, Li, Tian, Huang, Zhang, Wang, Chen, Du, Ge, Zhang, Pan, Wang, Chen, Jin, Chen, Lu, Zhou, Chen, Ye, Wang, Yu, Zhou, Pan, Li, Zhou, Wu, Yun, Pei, Sun, Wang, Zeng, Liu, Liang, Gao, Yu, Zhang, Xiao, An, Liu, Wang, Chen, Nie, Cheng, Liu, Xie, Liu, Yang, Li, Su, Lin, Li, Jin, Shen, Chen, Sun, Wang, Song, Zhou, Wang, Shan, Li, Wang, Wei, Zhang, Xu, Li, Zhao, Sun, Wang, Yu, Zhang, Shi, Xiong, He, Piao, Wang, Tan, Ma, Liu, Guo, Ou, Wang, Gong, Zou, He, Xiong, Luo, You, Liu, Zhou, Zhu, Huang, Li, Zheng, Zhu, Ma, Tang, Zha, Yan, Ren, Ren, Sha, Fu, Xu, Xie, Zhang, Hao, Ma, Yan, Wu, Gu, Zhu, Liu, Li, Xie, Song,
  Pan, Huang, Xu, Zhang, and Zhang]{DeepSeek-R1}
Daya Guo, Dejian Yang, Haowei Zhang, Junxiao Song, Peiyi Wang, Qihao Zhu, Runxin Xu, Ruoyu Zhang, Shirong Ma, Xiao Bi, Xiaokang Zhang, Xingkai Yu, Yu~Wu, Z.~F. Wu, Zhibin Gou, Zhihong Shao, Zhuoshu Li, Ziyi Gao, Aixin Liu, Bing Xue, Bingxuan Wang, Bochao Wu, Bei Feng, Chengda Lu, Chenggang Zhao, Chengqi Deng, Chong Ruan, Damai Dai, Deli Chen, Dongjie Ji, Erhang Li, Fangyun Lin, Fucong Dai, Fuli Luo, Guangbo Hao, Guanting Chen, Guowei Li, Hao Zhang, Hanwei Xu, Honghui Ding, Huazuo Gao, Hui Qu, Hui Li, Jianzhong Guo, Jiashi Li, Jingchang Chen, Jingyang Yuan, Jinhao Tu, Junjie Qiu, Junlong Li, J.~L. Cai, Jiaqi Ni, Jian Liang, Jin Chen, Kai Dong, Kai Hu, Kaichao You, Kaige Gao, Kang Guan, Kexin Huang, Kuai Yu, Lean Wang, Lecong Zhang, Liang Zhao, Litong Wang, Liyue Zhang, Lei Xu, Leyi Xia, Mingchuan Zhang, Minghua Zhang, Minghui Tang, Mingxu Zhou, Meng Li, Miaojun Wang, Mingming Li, Ning Tian, Panpan Huang, Peng Zhang, Qiancheng Wang, Qinyu Chen, Qiushi Du, Ruiqi Ge, Ruisong Zhang, Ruizhe Pan, Runji Wang, R.~J.
  Chen, R.~L. Jin, Ruyi Chen, Shanghao Lu, Shangyan Zhou, Shanhuang Chen, Shengfeng Ye, Shiyu Wang, Shuiping Yu, Shunfeng Zhou, Shuting Pan, S.~S. Li, Shuang Zhou, Shaoqing Wu, Tao Yun, Tian Pei, Tianyu Sun, Tao Wang, Wangding Zeng, Wen Liu, Wenfeng Liang, Wenjun Gao, Wenqin Yu, Wentao Zhang, W.~L. Xiao, Wei An, Xiaodong Liu, Xiaohan Wang, Xiaokang Chen, Xiaotao Nie, Xin Cheng, Xin Liu, Xin Xie, Xingchao Liu, Xinyu Yang, Xinyuan Li, Xuecheng Su, Xuheng Lin, X.~Q. Li, Xiangyue Jin, Xiaojin Shen, Xiaosha Chen, Xiaowen Sun, Xiaoxiang Wang, Xinnan Song, Xinyi Zhou, Xianzu Wang, Xinxia Shan, Y.~K. Li, Y.~Q. Wang, Y.~X. Wei, Yang Zhang, Yanhong Xu, Yao Li, Yao Zhao, Yaofeng Sun, Yaohui Wang, Yi~Yu, Yichao Zhang, Yifan Shi, Yiliang Xiong, Ying He, Yishi Piao, Yisong Wang, Yixuan Tan, Yiyang Ma, Yiyuan Liu, Yongqiang Guo, Yuan Ou, Yuduan Wang, Yue Gong, Yuheng Zou, Yujia He, Yunfan Xiong, Yuxiang Luo, Yuxiang You, Yuxuan Liu, Yuyang Zhou, Y.~X. Zhu, Yanping Huang, Yaohui Li, Yi~Zheng, Yuchen Zhu, Yunxian Ma, Ying
  Tang, Yukun Zha, Yuting Yan, Z.~Z. Ren, Zehui Ren, Zhangli Sha, Zhe Fu, Zhean Xu, Zhenda Xie, Zhengyan Zhang, Zhewen Hao, Zhicheng Ma, Zhigang Yan, Zhiyu Wu, Zihui Gu, Zijia Zhu, Zijun Liu, Zilin Li, Ziwei Xie, Ziyang Song, Zizheng Pan, Zhen Huang, Zhipeng Xu, Zhongyu Zhang, and Zhen Zhang.
\newblock Deepseek-r1 incentivizes reasoning in llms through reinforcement learning.
\newblock \emph{Nat.}, 645\penalty0 (8081):\penalty0 633--638, 2025.

\bibitem[Hu et~al.(2024)Hu, Wu, Wang, Xianyu, Zhang, and Cao]{OpenRLHF}
Jian Hu, Xibin Wu, Weixun Wang, Xianyu, Dehao Zhang, and Yu~Cao.
\newblock Openrlhf: An easy-to-use, scalable and high-performance {RLHF} framework.
\newblock \emph{CoRR}, abs/2405.11143, 2024.

\bibitem[Hu et~al.(2025)Hu, Zhang, Han, Jiang, Zhang, and Shum]{ORZ}
Jingcheng Hu, Yinmin Zhang, Qi~Han, Daxin Jiang, Xiangyu Zhang, and Heung{-}Yeung Shum.
\newblock Open-reasoner-zero: An open source approach to scaling up reinforcement learning on the base model.
\newblock In \emph{NeurIPS}, 2025.

\bibitem[Huang et~al.(2026)Huang, Liu, Tang, Lee, Yang, Chen, Yang, and Wu]{Math-Data2}
Hsiu{-}Yuan Huang, Weijie Liu, Chenming Tang, Sanwoo Lee, Kai Yang, Yangkun Chen, Saiyong Yang, and Yunfang Wu.
\newblock {RLVR} datasets and where to find them: Tracing data lineage for better training data.
\newblock \emph{CoRR}, abs/2605.26971, 2026.

\bibitem[Jiang et~al.(2024)Jiang, Lin, Zhong, Huang, Chen, Zhang, Peng, Li, Xie, Nong, Jia, He, Chen, Bai, Hou, Yan, Zhou, Sheng, Jiang, Xu, Wei, Zhang, Nie, Zou, Zhao, Xiang, Liu, Li, Jia, Ye, Jin, and Liu]{MegaScale}
Ziheng Jiang, Haibin Lin, Yinmin Zhong, Qi~Huang, Yangrui Chen, Zhi Zhang, Yanghua Peng, Xiang Li, Cong Xie, Shibiao Nong, Yulu Jia, Sun He, Hongmin Chen, Zhihao Bai, Qi~Hou, Shipeng Yan, Ding Zhou, Yiyao Sheng, Zhuo Jiang, Haohan Xu, Haoran Wei, Zhang Zhang, Pengfei Nie, Leqi Zou, Sida Zhao, Liang Xiang, Zherui Liu, Zhe Li, Xiaoying Jia, Jianxi Ye, Xin Jin, and Xin Liu.
\newblock Megascale: Scaling large language model training to more than 10, 000 gpus.
\newblock In \emph{{NSDI}}, pages 745--760. {USENIX} Association, 2024.

\bibitem[Kaplan et~al.(2020)Kaplan, McCandlish, Henighan, Brown, Chess, Child, Gray, Radford, Wu, and Amodei]{Scaling-law}
Jared Kaplan, Sam McCandlish, Tom Henighan, Tom~B. Brown, Benjamin Chess, Rewon Child, Scott Gray, Alec Radford, Jeffrey Wu, and Dario Amodei.
\newblock Scaling laws for neural language models.
\newblock \emph{CoRR}, abs/2001.08361, 2020.

\bibitem[Kimi(2026)]{Kimi-k2.5}
Kimi.
\newblock Kimi {K2.5:} visual agentic intelligence.
\newblock \emph{CoRR}, abs/2602.02276, 2026.

\bibitem[Kojima et~al.(2022)Kojima, Gu, Reid, Matsuo, and Iwasawa]{CoT-Kojima-NIPS-2022}
Takeshi Kojima, Shixiang~Shane Gu, Machel Reid, Yutaka Matsuo, and Yusuke Iwasawa.
\newblock Large language models are zero-shot reasoners.
\newblock In \emph{NeurIPS}, 2022.

\bibitem[Lightman et~al.(2024)Lightman, Kosaraju, Burda, Edwards, Baker, Lee, Leike, Schulman, Sutskever, and Cobbe]{Verify-Lightman-ICLR-2024}
Hunter Lightman, Vineet Kosaraju, Yuri Burda, Harrison Edwards, Bowen Baker, Teddy Lee, Jan Leike, John Schulman, Ilya Sutskever, and Karl Cobbe.
\newblock Let's verify step by step.
\newblock In \emph{{ICLR}}. OpenReview.net, 2024.

\bibitem[Liu et~al.(2023)Liu, Zaharia, and Abbeel]{Ring-attention}
Hao Liu, Matei Zaharia, and Pieter Abbeel.
\newblock Ring attention with blockwise transformers for near-infinite context.
\newblock \emph{CoRR}, abs/2310.01889, 2023.

\bibitem[Liu et~al.(2025{\natexlab{a}})Liu, Diao, Lu, Hu, Dong, Choi, Kautz, and Dong]{RLVR-discovery}
Mingjie Liu, Shizhe Diao, Ximing Lu, Jian Hu, Xin Dong, Yejin Choi, Jan Kautz, and Yi~Dong.
\newblock Prorl: Prolonged reinforcement learning expands reasoning boundaries in large language models.
\newblock \emph{CoRR}, abs/2505.24864, 2025{\natexlab{a}}.

\bibitem[Liu et~al.(2025{\natexlab{b}})Liu, Chen, Li, Qi, Pang, Du, Lee, and Lin]{DrGRPO}
Zichen Liu, Changyu Chen, Wenjun Li, Penghui Qi, Tianyu Pang, Chao Du, Wee~Sun Lee, and Min Lin.
\newblock Understanding r1-zero-like training: {A} critical perspective.
\newblock \emph{CoRR}, abs/2503.20783, 2025{\natexlab{b}}.

\bibitem[MiniMax(2025)]{CISPO}
MiniMax.
\newblock Minimax-m1: Scaling test-time compute efficiently with lightning attention.
\newblock \emph{CoRR}, abs/2506.13585, 2025.

\bibitem[Narayanan et~al.(2021)Narayanan, Shoeybi, Casper, LeGresley, Patwary, Korthikanti, Vainbrand, Kashinkunti, Bernauer, Catanzaro, Phanishayee, and Zaharia]{Megatron2}
Deepak Narayanan, Mohammad Shoeybi, Jared Casper, Patrick LeGresley, Mostofa Patwary, Vijay Korthikanti, Dmitri Vainbrand, Prethvi Kashinkunti, Julie Bernauer, Bryan Catanzaro, Amar Phanishayee, and Matei Zaharia.
\newblock Efficient large-scale language model training on {GPU} clusters using megatron-lm.
\newblock In \emph{{SC}}, page~58. {ACM}, 2021.

\bibitem[Qin et~al.(2024)Qin, Sun, Li, Shen, Sun, and Zhong]{lightning-attention}
Zhen Qin, Weigao Sun, Dong Li, Xuyang Shen, Weixuan Sun, and Yiran Zhong.
\newblock Lightning attention-2: {A} free lunch for handling unlimited sequence lengths in large language models.
\newblock \emph{CoRR}, abs/2401.04658, 2024.

\bibitem[Rajbhandari et~al.(2020)Rajbhandari, Rasley, Ruwase, and He]{Zero}
Samyam Rajbhandari, Jeff Rasley, Olatunji Ruwase, and Yuxiong He.
\newblock Zero: memory optimizations toward training trillion parameter models.
\newblock In \emph{{SC}}, page~20. {IEEE/ACM}, 2020.

\bibitem[Shao et~al.(2024)Shao, Wang, Zhu, Xu, Song, Zhang, Li, Wu, and Guo]{GRPO}
Zhihong Shao, Peiyi Wang, Qihao Zhu, Runxin Xu, Junxiao Song, Mingchuan Zhang, Y.~K. Li, Y.~Wu, and Daya Guo.
\newblock Deepseekmath: Pushing the limits of mathematical reasoning in open language models.
\newblock \emph{CoRR}, abs/2402.03300, 2024.

\bibitem[Sheng et~al.(2025)Sheng, Zhang, Ye, Wu, Zhang, Zhang, Peng, Lin, and Wu]{Verl}
Guangming Sheng, Chi Zhang, Zilingfeng Ye, Xibin Wu, Wang Zhang, Ru~Zhang, Yanghua Peng, Haibin Lin, and Chuan Wu.
\newblock Hybridflow: {A} flexible and efficient {RLHF} framework.
\newblock In \emph{EuroSys}, pages 1279--1297. {ACM}, 2025.

\bibitem[Shoeybi et~al.(2019)Shoeybi, Patwary, Puri, LeGresley, Casper, and Catanzaro]{Megatron1}
Mohammad Shoeybi, Mostofa Patwary, Raul Puri, Patrick LeGresley, Jared Casper, and Bryan Catanzaro.
\newblock Megatron-lm: Training multi-billion parameter language models using model parallelism.
\newblock \emph{CoRR}, abs/1909.08053, 2019.

\bibitem[Tang et~al.(2025)Tang, Zhan, Li, Zhao, Zhang, Wen, Zhang, and Zhou]{A3PO}
Xinyu Tang, Yuliang Zhan, Zhixun Li, Wayne~Xin Zhao, Zhenduo Zhang, Zujie Wen, Zhiqiang Zhang, and Jun Zhou.
\newblock Rethinking sample polarity in reinforcement learning with verifiable rewards.
\newblock \emph{CoRR}, abs/2512.21625, 2025.

\bibitem[Team and AI(2025)]{Ling-1T}
Ling Team and Inclusion AI.
\newblock Every step evolves: Scaling reinforcement learning for trillion-scale thinking model.
\newblock \emph{CoRR}, abs/2510.18855, 2025.

\bibitem[Wang et~al.(2025)Wang, Niu, Gao, Zheng, Xu, Meng, Li, Liu, Chen, Zhu, Wu, Wang, and Chen]{parallel-reason1}
Ziqi Wang, Boye Niu, Zipeng Gao, Zhi Zheng, Tong Xu, Linghui Meng, Zhongli Li, Jing Liu, Yilong Chen, Chen Zhu, Hua Wu, Haifeng Wang, and Enhong Chen.
\newblock A survey on parallel reasoning.
\newblock \emph{CoRR}, abs/2510.12164, 2025.

\bibitem[Wei et~al.(2022)Wei, Wang, Schuurmans, Bosma, Ichter, Xia, Chi, Le, and Zhou]{CoT-Wei-NIPS-2022}
Jason Wei, Xuezhi Wang, Dale Schuurmans, Maarten Bosma, Brian Ichter, Fei Xia, Ed~H. Chi, Quoc~V. Le, and Denny Zhou.
\newblock Chain-of-thought prompting elicits reasoning in large language models.
\newblock In \emph{NeurIPS}, 2022.

\bibitem[Yao et~al.(2025)Yao, Liu, Zhang, Dong, Shang, and Gao]{TIS}
Feng Yao, Liyuan Liu, Dinghuai Zhang, Chengyu Dong, Jingbo Shang, and Jianfeng Gao.
\newblock Your efficient rl framework secretly brings you off-policy rl training, August 2025.
\newblock \url{https://fengyao.notion.site/off-policy-rl}.

\bibitem[Yu et~al.(2025)Yu, Zhang, Zhu, Yuan, Zuo, Yue, Fan, Liu, Liu, Liu, Lin, Lin, Ma, Sheng, Tong, Zhang, Zhang, Zhang, Zhu, Zhu, Chen, Chen, Wang, Yu, Dai, Song, Wei, Zhou, Liu, Ma, Zhang, Yan, Qiao, Wu, and Wang]{DAPO}
Qiying Yu, Zheng Zhang, Ruofei Zhu, Yufeng Yuan, Xiaochen Zuo, Yu~Yue, Tiantian Fan, Gaohong Liu, Lingjun Liu, Xin Liu, Haibin Lin, Zhiqi Lin, Bole Ma, Guangming Sheng, Yuxuan Tong, Chi Zhang, Mofan Zhang, Wang Zhang, Hang Zhu, Jinhua Zhu, Jiaze Chen, Jiangjie Chen, Chengyi Wang, Hongli Yu, Weinan Dai, Yuxuan Song, Xiangpeng Wei, Hao Zhou, Jingjing Liu, Wei{-}Ying Ma, Ya{-}Qin Zhang, Lin Yan, Mu~Qiao, Yonghui Wu, and Mingxuan Wang.
\newblock {DAPO:} an open-source {LLM} reinforcement learning system at scale.
\newblock \emph{CoRR}, abs/2503.14476, 2025.

\bibitem[Yue et~al.(2025{\natexlab{a}})Yue, Chen, Lu, Zhao, Wang, Yue, Song, and Huang]{RLVR-sharpen}
Yang Yue, Zhiqi Chen, Rui Lu, Andrew Zhao, Zhaokai Wang, Yang Yue, Shiji Song, and Gao Huang.
\newblock Does reinforcement learning really incentivize reasoning capacity in llms beyond the base model?
\newblock \emph{CoRR}, abs/2504.13837, 2025{\natexlab{a}}.

\bibitem[Yue et~al.(2025{\natexlab{b}})Yue, Yuan, Yu, Zuo, Zhu, Xu, Chen, Wang, Fan, Du, Wei, Yu, Liu, Liu, Liu, Lin, Lin, Ma, Zhang, Zhang, Zhang, Zhu, Zhang, Liu, Wang, Wu, and Yan]{VAPO}
Yu~Yue, Yufeng Yuan, Qiying Yu, Xiaochen Zuo, Ruofei Zhu, Wenyuan Xu, Jiaze Chen, Cheng{-}Xiang Wang, Tiantian Fan, Zhengyin Du, Xiangpeng Wei, Xiangyu Yu, Gaohong Liu, Juncai Liu, Lingjun Liu, Haibin Lin, Zhiqi Lin, Bole Ma, Chi Zhang, Mofan Zhang, Wang Zhang, Hang Zhu, Ru~Zhang, Xin Liu, Mingxuan Wang, Yonghui Wu, and Lin Yan.
\newblock {VAPO:} efficient and reliable reinforcement learning for advanced reasoning tasks.
\newblock \emph{CoRR}, abs/2504.05118, 2025{\natexlab{b}}.

\bibitem[Zeng et~al.(2025{\natexlab{a}})Zeng, Huang, Liu, Liu, He, Ma, and He]{Math-Data3}
Weihao Zeng, Yuzhen Huang, Qian Liu, Wei Liu, Keqing He, Zejun Ma, and Junxian He.
\newblock Simplerl-zoo: Investigating and taming zero reinforcement learning for open base models in the wild.
\newblock \emph{CoRR}, abs/2503.18892, 2025{\natexlab{a}}.

\bibitem[Zeng et~al.(2025{\natexlab{b}})Zeng, Huang, Liu, Liu, He, Ma, and He]{SimpleRL-zoo}
Weihao Zeng, Yuzhen Huang, Qian Liu, Wei Liu, Keqing He, Zejun Ma, and Junxian He.
\newblock Simplerl-zoo: Investigating and taming zero reinforcement learning for open base models in the wild.
\newblock \emph{CoRR}, abs/2503.18892, 2025{\natexlab{b}}.

\bibitem[Zhao et~al.(2026)Zhao, Zhou, Li, Tang, Dong, Hou, Zhang, Min, Zhang, Liu, Wang, Du, Yang, Chen, Chen, Jiang, Ren, Li, Tang, Liu, Hu, Nie, and Wen]{LLM-survey}
Wayne~Xin Zhao, Kun Zhou, Junyi Li, Tianyi Tang, Zican Dong, Yupeng Hou, Beichen Zhang, Yingqian Min, Junjie Zhang, Peiyu Liu, Xiaolei Wang, Yifan Du, Chen Yang, Yushuo Chen, Zhipeng Chen, Jinhao Jiang, Ruiyang Ren, Yifan Li, Xinyu Tang, Zikang Liu, Yiwen Hu, Jian{-}Yun Nie, and Ji{-}Rong Wen.
\newblock A survey of large language models.
\newblock \emph{Frontiers Comput. Sci.}, 20\penalty0 (12):\penalty0 2012627, 2026.

\bibitem[Zheng et~al.(2025{\natexlab{a}})Zheng, Liu, Li, Chen, Yu, Gao, Dang, Liu, Men, Yang, Zhou, and Lin]{GSPO}
Chujie Zheng, Shixuan Liu, Mingze Li, Xiong{-}Hui Chen, Bowen Yu, Chang Gao, Kai Dang, Yuqiong Liu, Rui Men, An~Yang, Jingren Zhou, and Junyang Lin.
\newblock Group sequence policy optimization.
\newblock \emph{CoRR}, abs/2507.18071, 2025{\natexlab{a}}.

\bibitem[Zheng et~al.(2025{\natexlab{b}})Zheng, Zhang, Yu, Wang, Dai, Liu, Bao, Huang, Huang, and Yu]{parallel-reason2}
Tong Zheng, Hongming Zhang, Wenhao Yu, Xiaoyang Wang, Runpeng Dai, Rui Liu, Huiwen Bao, Chengsong Huang, Heng Huang, and Dong Yu.
\newblock Parallel-r1: Towards parallel thinking via reinforcement learning.
\newblock \emph{CoRR}, abs/2509.07980, 2025{\natexlab{b}}.

\bibitem[Zhu et~al.(2025)Zhu, Xie, Lv, and slime Contributors]{slime}
Zilin Zhu, Chengxing Xie, Xin Lv, and slime Contributors.
\newblock slime: An llm post-training framework for rl scaling.
\newblock \url{https://github.com/THUDM/slime}, 2025.
\newblock GitHub repository. Corresponding author: Xin Lv.

\end{thebibliography}

\newpage
\appendix

\section{Comparison of Our CoT with Other Models}

In this section, we compare the Chain-of-Thought~(CoT) reasoning traces generated by our model against those produced by other frontier models. 
As presented below, our model produces reasoning traces that exhibit high token efficiency and a much clearer, highly structured format.

\begin{promptbox}[Question]{black}
Consider a sequence of positive integers with a total sum of $2019$ such that no number and no sum of a set of consecutive numbers is equal to $40$. What is the greatest possible length of such a sequence?

Please reason step by step, and put your final answer within boxed \{\}.
\end{promptbox}

\begin{promptbox}[Ring-Zero (3301 tokens)]{teal}
\textbf{Step 1: Reformulation.} Define prefix sums $S_0=0, S_1, \ldots, S_n=2019$. The forbidden condition $S_j - S_i \neq 40$ means we select a subset of $\{0,\ldots,2019\}$ containing $0$ and $2019$ with no two elements at distance 40.

\textbf{Step 2: Counting by residue mod 40.} Since $2019 = 50\times 40 + 19$: residues $0$--$19$ yield chains of 51 elements; residues $20$--$39$ yield chains of 50. From each chain of $k$ elements, at most $\lceil k/2\rceil$ can be chosen with no two adjacent.

$\ldots$

\textbf{Step 5: Achievability of the bound.} For chain $r=0$ (containing $0$): selecting even-indexed positions $\{0, 80, 160, \ldots, 2000\}$ gives 26 elements including $0$. For chain $r=19$ (containing $2019$): selecting $\{19, 99, 179, \ldots, 2019\}$ gives 26 elements including $2019$. All other chains are unconstrained.

\textbf{Step 6: Verification.} Total $= 20\times 26 + 20\times 25 = 1020$ prefix sums. The sequence has $n = 1019$ terms, each $a_k = S_k - S_{k-1} \ge 1$ and $\neq 40$ (since no two $S$-values differ by 40). All subarray sums $\neq 40$ by construction.

Answer: $\boxed{1019}$.
\end{promptbox}

\begin{promptbox}[Claude Opus 4.7 (Summary: 790 tokens)]{violet}
We seek the longest sequence of positive integers $a_1, a_2, \ldots, a_n$ with $\sum a_i = 2019$, subject to the constraint that no single term and no sum of consecutive terms equals $40$.

\textbf{Reformulation via Partial Sums.}
Define $S_0 = 0$ and $S_k = a_1 + a_2 + \cdots + a_k$, so $S_n = 2019$. The constraint becomes: $S_j - S_i \neq 40$ for all $0 \le i < j \le n$.

Thus we must choose the largest possible subset of $\{0, 1, 2, \ldots, 2019\}$ that contains both $0$ and $2019$, with no two elements differing by exactly $40$.

\textbf{Graph-Theoretic Structure.}
Consider the graph on $\{0, 1, \ldots, 2019\}$ where vertices $x$ and $y$ are adjacent iff $|x - y| = 40$. This graph decomposes into 40 disjoint paths, one per residue class modulo $40$.

$\ldots$

\textbf{Maximum Independent Set.}
\begin{itemize}
\item Chains of length 51: $\lceil 51/2 \rceil = 26$
\item Chains of length 50: $\lceil 50/2 \rceil = 25$
\end{itemize}

Maximum subset size: $20 \cdot 26 + 20 \cdot 25 = 520 + 500 = 1020$.

Both mandatory vertices ($0$ and $2019$) are endpoints at even positions in odd-length chains, so inclusion costs nothing.

The maximum subset size is $1020$, giving a sequence of length $n = 1020 - 1 = \boxed{1019}$.
\end{promptbox}

\begin{promptbox}[GPT-5.5 (Summary: 522 tokens)]{blue}
Let the sequence have length $n$, and define its prefix sums
$p_0=0,\; p_i=a_1+a_2+\cdots+a_i \;(1\le i\le n).$
Then $0=p_0<p_1<\cdots<p_n=2019.$

A sum of consecutive terms equals $40$ exactly when two prefix sums differ by $40$. So we need a set of prefix sums in $\{0,1,\dots,2019\}$ with no two differing by $40$.

Now group the integers $0,1,\dots,2019$ by their residue modulo $40$. In each residue class, the numbers form a chain $r,\; r+40,\; r+80,\dots$ and we may not choose two adjacent numbers from this chain.

Since $2019=40\cdot 50+19$, the residue classes $0,1,\dots,19$ each contain $51$ numbers, and the residue classes $20,21,\dots,39$ each contain $50$ numbers.

From a chain of length $m$, the maximum number of chosen elements with no adjacent pair is $\lceil m/2\rceil$. Therefore the total number of possible prefix sums is at most
$20\left\lceil \frac{51}{2}\right\rceil+20\left\lceil \frac{50}{2}\right\rceil = 20\cdot 26+20\cdot 25 = 1020.$

$\ldots$

It remains to show this is attainable. Consider the sequence consisting of $25$ blocks of $\underbrace{1,1,\dots,1}_{39\text{ ones}},\; 41$ followed by $19$ more ones. Its length is $25\cdot 40+19=1019$, and its sum is $25(39+41)+19=2019$.

Therefore the greatest possible length is $\boxed{1019}$.
\end{promptbox}

\begin{promptbox}[Gemini 3.1 Pro (Summary: 1014 tokens)]{red}
Let the sequence be $a_1, a_2, \dots, a_n$. Define prefix sums $S_0 = 0$, $S_k = a_1 + \dots + a_k$. The condition that no consecutive sum equals $40$ means $S_j - S_i \neq 40$ for any $0 \le i < j \le n$.

\textbf{Grouping by Residue Modulo 40.} Divide $\{0, \ldots, 2019\}$ into $40$ residue classes. Since $2019 = 40 \times 50 + 19$:
\begin{itemize}
\item Residues $r \in \{0, \dots, 19\}$: each class contains $51$ elements.
\item Residues $r \in \{20, \dots, 39\}$: each class contains $50$ elements.
\end{itemize}

$\ldots$

\textbf{Maximizing the Selection.} For a path of $m$ elements, max independent set size is $\lceil m/2 \rceil$.

For $r = 0$: must include $S_0 = 0$, can achieve $26$. For $r = 19$: must include $S_n = 2019$, can achieve $26$.

\textbf{Total:} $\text{Max } |S| = 20 \times 26 + 20 \times 25 = 1020$.

Since $|S| = n + 1$: $n + 1 = 1020 \implies n = \boxed{1019}$.
\end{promptbox}

\begin{promptbox}[DeepSeek-V4-Pro (921 tokens)]{orange}
\textbf{Step 1: Understanding the problem.} We have a sequence of positive integers with total sum $2019$, no number equals $40$, no sum of any consecutive block equals $40$. We want to maximize the length.

\textbf{Step 2:} Define partial sums $S_0=0 < S_1 < \cdots < S_n = 2019$. The constraint ``no consecutive sub-sum equals $40$'' becomes: no two partial sums differ by exactly $40$.

\textbf{Step 3:} We need the largest subset of $\{0,1,\dots,2019\}$ containing $0$ and $2019$ with no two elements differing by $40$.

$\ldots$

\textbf{Step 4:} Partition $\{0,\dots,2019\}$ by residue mod $40$. Classes $r=0,\dots,19$ have $51$ elements; classes $r=20,\dots,39$ have $50$ elements. Within each class, elements form a path graph where adjacent vertices differ by $40$.

\textbf{Step 5:} Maximum independent set of a path $P_k$ has size $\lceil k/2\rceil$. For $k=51$: $26$; for $k=50$: $25$. Element $0$ lies in class $r=0$ (achievable at max), $2019$ in class $r=19$ (achievable at max).

\textbf{Step 6:} Total $= 20\times 26 + 20\times 25 = 1020$ partial sums. Sequence length $= 1020 - 1 = 1019$.

\textbf{Final answer:} $\boxed{1019}$
\end{promptbox}

\begin{promptbox}[GLM 5.1 (39192 tokens)]{cyan}
\textit{(Very long reasoning process with multiple false starts and corrections.)}

\textbf{Step 1:} To maximize length, use smallest integers. Cannot have 40 consecutive 1s. Insert a larger number after every block of 39 ones.

\textbf{Step 2--3:} Try pattern $(1^{39}, 2)$: sum per block $= 41$, length $= 40$. But $38 + 2 = 40$, forbidden!

\textbf{Correction:} Try $(1^{39}, 3)$: still $37 + 3 = 40$, forbidden.

$\ldots$

\textbf{Final correct approach:} Any separator $c_j$ must satisfy $c_j \ge 41$ (since $k$ ones before plus $c_j$ reaching 40 is forbidden for any $c_j < 41$).

\textbf{Constraint:} $39(k+1) + 41k \ge 2019 \implies 80k + 39 \ge 2019 \implies k \ge 24.75$, so $k = 25$.

With $k=25$: sum of ones $= 994$, distributed into 26 gaps each $\le 39$.

Length $= 994 + 25 = \boxed{1019}$.
\end{promptbox}

\begin{promptbox}[Kimi K2.6 (4026 tokens)]{purple}
\textbf{Strategy: Using 1s and One Larger Number.}

Use blocks of 39 ones separated by a larger number. If $x = 41$, then any sum involving $x$ is at least 41, so no sum equals 40.

Pattern: $(1^{39}, 41)^k$ -- each block is 40 numbers summing to 80.

With $k = 25$: sum $= 2000$, using 1000 numbers. Remaining $19$ ones appended safely.

$\ldots$

\textbf{Can we get 1020 or more?} With 999 twos and 21 ones (length 1020, sum 2019): twenty consecutive 2s $= 40$, forbidden. By pigeonhole, some run has $\ge 46$ twos. Fails.

\textbf{Optimization:} Using $41$ as separator with $k$ separators and $(k+1)$ gaps of $\le 39$ ones:

Length $= 2019 - 40k$. Need $2019 - 41k \leq 39(k+1)$, giving $k \geq 24.75$, so $k = 25$.

Maximum length $= 2019 - 40 \times 25 = \boxed{1019}$.
\end{promptbox}

\begin{promptbox}[Qwen 3.5 (6807 tokens)]{olive}
\textbf{Initial attempts:} Cannot have 40 consecutive 1s. Try pattern $39$ ones then $2$: but $38 + 2 = 40$, forbidden. Any $x < 40$ fails since $(40-x)$ ones $+ x = 40$.

So any inserted number between blocks of 1s must be $\ge 41$; the minimum is $41$.

$\ldots$

\textbf{Graph Theory Interpretation:} The set of prefix sums $\{S_0, S_1, \dots, S_n\} \subset \{0,\ldots,2019\}$ must have no two elements differing by exactly 40. This is a maximum independent set problem.

\textbf{Chain Analysis:} $2019 = 50 \times 40 + 19$.
\begin{itemize}
\item Residues $0$--$19$: 51 vertices each, max IS $= 26$.
\item Residues $20$--$39$: 50 vertices each, max IS $= 25$.
\end{itemize}

Total: $20 \times 26 + 20 \times 25 = 1020$.

The greatest possible length is $1020 - 1 = \boxed{1019}$.
\end{promptbox}

\begin{promptbox}[Doubao 2.0 (5122 tokens)]{brown}
\textit{(Includes extended thinking process.)}

\textbf{Key Observations:}
\begin{enumerate}
\item To maximize length, use the smallest positive integer $1$ as much as possible.
\item Any non-1 number $x < 40$ is forbidden: $(40-x)$ consecutive 1s plus $x$ sums to $40$.
\item $x=40$ itself is forbidden. So the smallest allowed separator is $41$.
\end{enumerate}

\textbf{Setup:} Let $n$ = number of 41s, $m$ = number of 1s. Total sum: $m + 41n = 2019$. Total length $L = m + n = 2019 - 40n$. To maximize $L$, minimize $n$.

$\ldots$

\textbf{Constraint:} With $n$ separators of $41$, the 1s are split into $(n+1)$ gaps, each $\le 39$.

$2019 - 41n \leq 39(n+1) \implies 1980 \leq 80n \implies n \geq 24.75$

Minimum $n = 25$. Maximum length: $L = 2019 - 40(25) = \boxed{1019}$.
\end{promptbox}

\begin{promptbox}[MiniMax 2.7 (23201 tokens)]{magenta}
\textit{(Extended thinking with detailed exploration of multiple approaches.)}

\textbf{Key insight:} Use numbers $>40$ as separators. Any subarray crossing such a number has sum $>40$, so cannot equal 40. Within each block, numbers are from $1$ to $39$.

\textbf{Strategy:} Use many 1s in blocks separated by 41s. A block of $\le 39$ ones has max subarray sum $39 < 40$.

$\ldots$

\textbf{Optimization:} With $k$ separators of value 41 and blocks of $\le 39$ ones:
\begin{itemize}
\item Sum: $\sum r_i + 41k = 2019$, where $r_i \le 39$
\item Length: $\sum r_i + k = 2019 - 40k$
\end{itemize}

Need $2019 - 41k \le 39(k+1)$, giving $k \ge 25$.

\textbf{Construction:} $\underbrace{39+1+39+1+\cdots+39+1}_{25\text{ full blocks}}+\underbrace{19}_{\text{last partial block}}=1019$ terms.

Sum: $975\cdot1+25\cdot41+19\cdot1 = 2019$. Answer: $\boxed{1019}$.
\end{promptbox}


\section{LLM-as-a-Judge Evaluation Prompts}
\label{sec:eval_prompts}

Assessing the intrinsic quality of the underlying reasoning process requires a deep logical review. To achieve this at scale, we employ an \emph{LLM-as-a-Judge} protocol for evaluations. This automated judge strictly examines the reasoning traces based on three core criteria. These criteria are logical coherence, causal explicitness, and the absence of hallucinations.
Specifically, we conduct pairwise comparisons between the Chain-of-Thought (CoT) traces of \OURS and four state-of-the-art baselines (\ie GLM-5.1, Kimi-k2.6, MiniMax-2.7, and Qwen3.5-397B). We perform the evaluations across the AIME 2024, AIME 2025, and AIME 2026 benchmarks. For each match, the judge receives a single math problem alongside two competing reasoning traces. We randomly shuffle the presentation order of Model A and Model B to strictly eliminate position bias. The judge then delivers a structured verdict of \texttt{A}, \texttt{Tie}, or \texttt{B}.

\begin{promptbox}[Comprehensibility Evaluation Prompt]{black}
You are an expert evaluator assessing the \textbf{comprehensibility} of mathematical reasoning traces (Chain-of-Thought). You will be given a math problem and two reasoning traces generated by different models (Model~A and Model~B). Your task is to judge which trace is more comprehensible to a human reader.

\textbf{Evaluation Criteria.} Comprehensibility measures the degree to which a human reader can follow and fully understand the reasoning steps without requiring external context or repeated reading. Evaluate along three dimensions:

\begin{enumerate}
\item \textbf{Logical Coherence:} Does each reasoning step follow naturally from the preceding one? Is there a clear, smooth progression from the problem statement toward the conclusion? Are there any abrupt jumps, non-sequiturs, or unexplained transitions?

\item \textbf{Causal Explicitness:} Are the causal dependencies between steps explicitly stated? Can the reader understand not only \emph{what} the model concludes at each step, but also \emph{why} each intermediate result is derived? Are key logical connections made explicit rather than left implicit?

\item \textbf{Absence of Hallucinations:} Is the content free from hallucinated facts, incorrect intermediate results, or unsupported assertions? Are all mathematical claims justified or derivable from prior steps?
\end{enumerate}

\textbf{Scoring Rubric.}
\begin{itemize}
\item \texttt{A}: Model~A's trace is more comprehensible---it is clearer, better structured, or more reliable than Model~B's.
\item \texttt{Tie}: Both traces are roughly comparable in comprehensibility---neither has a meaningful advantage.
\item \texttt{B}: Model~B's trace is more comprehensible---it is clearer, better structured, or more reliable than Model~A's.
\end{itemize}

\textbf{Important Notes.}
\begin{itemize}
\item Focus \textbf{only} on comprehensibility of the reasoning process, \textbf{not} on whether the final answer is correct.
\item A shorter trace is not inherently better or worse---judge by clarity, not length.
\item Consider the perspective of a mathematically literate reader.
\item If both traces contain errors, judge which one is still easier to follow and less misleading overall.
\end{itemize}

\textbf{Input:} [Problem], [Model~A's Reasoning Trace], [Model~B's Reasoning Trace].

\textbf{Output Format:}
\begin{verbatim}
{
  "analysis": "<2-4 sentence comparative analysis>",
  "verdict": "<A | Tie | B>"
}
\end{verbatim}
\end{promptbox}
\section{Showcase of Our Reasoning Traces}
\label{app:showcase}

In this section, we present a showcase of Chain-of-Thought (CoT) reasoning traces generated by \OURS. 
As demonstrated below, our model autonomously produces highly structured and readable derivations without the need for external formatting supervision.

\begin{promptbox}[AIME 2026]{teal}
\textbf{Question.} Find the number of ordered 7-tuples $(a_1, \ldots, a_7)$ with $a_k\in\{1,2,3\}$ such that $a_1+\cdots+a_7\equiv 0\pmod{3}$ and $a_1a_2a_4 + a_2a_3a_5 + a_3a_4a_6 + a_4a_5a_7 + a_5a_6a_1 + a_6a_7a_2 + a_7a_1a_3\equiv 0\pmod{3}$.

\textbf{Step 1: Reduction mod 3.} Since $a_k\in\{1,2,3\}$, both conditions depend only on residues mod~3. Let $b_k = a_k\bmod 3\in\{1,2,0\}$. Define $S(\mathbf{b})=\sum_{k=1}^{7}b_k$ and the cyclic product polynomial $P(\mathbf{b})=\sum_{k=1}^{7}b_k\, b_{k+1}\, b_{k+3}$ with indices mod~7. We need both $S\equiv 0$ and $P\equiv 0$ modulo~3.

\textbf{Step 2: Double character sum filter.} Apply a discrete Fourier filter with $\omega=e^{2\pi i/3}$:
\[N = \tfrac{1}{9}\textstyle\sum_{j=0}^{2}\sum_{l=0}^{2}\Sigma(j,l),\quad \Sigma(j,l) = \sum_{\mathbf{b}\in\{0,1,2\}^7}\omega^{j\cdot S(\mathbf{b})+l\cdot P(\mathbf{b})}.\]
Clearly $\Sigma(0,0)=3^7=2187$. For $l=0,\,j\ne 0$: $\Sigma(j,0)=\bigl(\sum_{b=0}^{2}\omega^{jb}\bigr)^{7}=0$.

\textbf{Step 3: Symmetry reductions.} The bijection $b_k\mapsto 2b_k\pmod{3}$ sends $S\to 2S$ and $P\to 8P\equiv 2P$, giving $\Sigma(j,l)=\Sigma(2j,2l)$ mod~3. Combining with complex conjugation $\overline{\Sigma(j,l)}=\Sigma(-j,-l)$, we obtain $\Sigma(0,1)=\Sigma(0,2)\in\mathbb{R}$, $\Sigma(1,1)=\Sigma(2,2)\in\mathbb{R}$, and $\Sigma(1,2)=\Sigma(2,1)\in\mathbb{R}$, so only three values remain to compute.

\textbf{Step 4: Evaluating $\Sigma(0,1)$.} Let $n_0,n_1,n_2$ count tuples with $P\equiv 0,1,2$. The $b_k\mapsto 2b_k$ symmetry maps $P\equiv 1$ to $P\equiv 2$, forcing $n_1=n_2$. From $n_0+2n_1=2187$ and $\Sigma(0,1)=n_0-n_1$, direct enumeration of the $\mathbb{Z}_7$-cyclic orbits of $P$ gives $n_0=1107$, $n_1=n_2=540$, so $\Sigma(0,1)=567$.

\textbf{Step 5: Evaluating $\Sigma(1,1)$ and $\Sigma(1,2)$.} These couple $S$ and $P$, breaking the product structure. Exploiting the $\mathbb{Z}_7$-cyclic symmetry of the index pattern $(k,k{+}1,k{+}3)$, decompose the sum over orbits of the cyclic shift. Partitioning the $3^7$ tuples by the number of zero entries and enumerating the resulting character sums yields $\Sigma(1,1)=27$ and $\Sigma(1,2)=81$.

\textbf{Step 6: Final assembly.}
$N = \tfrac{1}{9}\bigl[2187 + 2(567)+2(27)+2(81)\bigr] = \tfrac{1}{9}\times 3537 = \boxed{393}.$
\end{promptbox}

\vspace{0.5em}

\begin{promptbox}[HMMT Feb 2026]{teal}
\textbf{Question.} Let $A_1, A_2, \ldots$ be finite nonempty sets of positive integers with $|A_i \cap A_j| = \gcd(i,j)$ for all $i,j$. Compute the minimum of $\sum_{d\mid 250}\max A_d$.

\textbf{Step 1: Inclusion lattice.} Setting $i=j$ gives $|A_i|=i$. For $d\mid n$, $|A_d\cap A_n|=d=|A_d|$, so $A_d\subseteq A_n$. The family $(A_d)_{d\mid 250}$ forms an inclusion poset mirroring the divisibility lattice of $250$.

\textbf{Step 2: M\"obius decomposition.} Define atoms $B_d = A_d\setminus\bigcup_{d'\mid d,\,d'<d}A_{d'}$. By M\"obius inversion on the divisor lattice, $|B_d|=\varphi(d)$ and $A_n = \bigsqcup_{d\mid n}B_d$. For $250=2\cdot 5^3$, the 8 divisors have atom sizes:
\begin{center}\small
\begin{tabular}{c|cccccccc}
$d$ & 1 & 2 & 5 & 10 & 25 & 50 & 125 & 250\\
\hline
$\varphi(d)$ & 1 & 1 & 4 & 4 & 20 & 20 & 100 & 100
\end{tabular}
\end{center}
The total $\sum\varphi(d)=250$, so exactly the integers $\{1,\ldots,250\}$ are distributed among the atoms.

\textbf{Step 3: Optimization structure.} Since $\max A_d = \max_{d'\mid d}\max B_{d'}$, the objective depends on which integers each atom receives. Assigning each atom a contiguous block is optimal: it minimizes the max of every atom for its ordinal position. The problem reduces to choosing the \emph{filling order} of the 8 atoms.

\textbf{Step 4: Greedy ordering.} Atoms appearing in more sets $A_d$ should receive smaller integers, since their max propagates to more terms in the sum. The atom $B_1$ appears in all 8 sets~$A_d$; $B_5$ in 6; $B_2$ and $B_{25}$ each in 4; $B_{10}$ in 3; $B_{50}$ and $B_{125}$ each in 2; $B_{250}$ in 1. Breaking ties by size (smaller first), the optimal order is $B_1, B_2, B_5, B_{10}, B_{25}, B_{50}, B_{125}, B_{250}$, yielding cumulative maxima $1,2,6,10,30,50,150,250$.

\textbf{Step 5: Verification and result.} Under this assignment: $\max A_1=1$, $\max A_2=\max(1,2)=2$, $\max A_5=\max(1,6)=6$, $\max A_{10}=\max(1,2,6,10)=10$, $\max A_{25}=\max(1,6,30)=30$, $\max A_{50}=50$, $\max A_{125}=\max(1,6,30,150)=150$, $\max A_{250}=250$.

$\sum_{d\mid 250}\max A_d = 1+2+6+10+30+50+150+250 = \boxed{499}.$
\end{promptbox}

\vspace{0.5em}

\begin{promptbox}[IMO AnswerBench]{teal}
\textbf{Question.} In a marketplace, $7396$ stalls are in a line. Two merchants each sell $k$ items with increasing sell/buy stall numbers. Two stalls are \emph{connected} by a merchant if one can travel between them via buy-sell chains. Determine the smallest $k$ guaranteeing two stalls connected by both merchants.

\textbf{Step 1: Graph-theoretic model.} Each merchant's $k$ transactions create $k$ edges on the $n=7396$ stalls. The ``increasing'' constraint forces the transaction graph to be a forest, giving $c=n-k$ connected components (each a contiguous interval of stalls).

\textbf{Step 2: Orthogonality criterion.} Two stalls share a connection from both merchants iff they lie in the same component of \emph{both} partitions. Call two interval partitions \emph{orthogonal} if every pair of components (one from each merchant) shares at most one stall. Orthogonality requires $n\leq c_A\cdot c_B$. With both merchants using $k$ edges, $c_A=c_B=n-k$, so orthogonality demands $(n-k)^2\geq n$.

\textbf{Step 3: Threshold from $n=86^2$.} Since $n=7396=86^2$, the critical boundary is $n-k=86$, i.e., $k=7310$. For $k\leq 7310$: $(n-k)^2\geq 86^2=7396=n$, so an orthogonal pair exists. For $k=7311$: $c=85$ and $85^2=7225<7396$, so orthogonality is impossible by the pigeonhole principle---some two stalls must share a component from both merchants.

\textbf{Step 4: Explicit construction for $k=7310$.} Arrange the 7396 stalls as an $86\times 86$ grid (row-major order). Merchant~A connects consecutive stalls within each row, forming 86 row-components of size~86; Merchant~B connects stalls within each column, forming 86 column-components. Each (row, column) pair meets in exactly one stall, so the two partitions are orthogonal. Each merchant uses $86\times 85=7310$ edges.

\textbf{Step 5: Impossibility for $k=7311$.} With $c=85$ components per merchant, any two partitions can produce at most $85\times 85=7225$ distinct component pairs. Since $7225<7396$, at least two stalls must receive the same pair of labels, placing them in the same component of both merchants.

Answer: $\boxed{7311}$.
\end{promptbox}

\end{document}